\documentclass[preprint]{elsarticle}

\usepackage{lineno,hyperref}
\modulolinenumbers[5]

\usepackage{amsfonts}
\usepackage{booktabs}
\usepackage{graphicx}
\usepackage{wrapfig}
\usepackage{amsmath}
\usepackage{amssymb}
\usepackage{subfigure}
\usepackage{longtable}
\usepackage{booktabs}
\usepackage{multirow}

\DeclareMathAlphabet\mathbfcal{OMS}{cmsy}{b}{n}
\newcommand{\fb}[1]{\dofb#1}

\journal{Neural Networks}
\newcommand{\dofb}[1]{\textbf{#1}\nobreak\hspace{0pt}}
\begin{document}

\begin{frontmatter}

\title{How to train your draGAN: A task oriented solution to imbalanced classification}

\author{Leon O. Guertler\corref{mycorrespondingauthor}}
\ead{leon002@e.nut.edu.sg}

\author{Andri Ashfahani}
\author{Anh Tuan Luu}

\cortext[mycorrespondingauthor]{Corresponding author}

\begin{abstract}
    The long-standing challenge of building effective classification models for small and imbalanced datasets has seen little improvement since the creation of the Synthetic Minority Over-sampling Technique (SMOTE) over 20 years ago. Though GAN based models seem promising, there has been a lack of 
    purpose built architectures for solving the aforementioned problem, as most previous studies focus on applying already existing models. This paper proposes a unique, performance-oriented, data-generating strategy that utilizes a new architecture, coined draGAN, to generate both minority and majority samples. The samples are generated with the objective of optimizing the classification model’s performance, rather than similarity to the real data. We benchmark our approach against state-of-the-art methods from the SMOTE family and competitive GAN based approaches on 94 tabular datasets with varying degrees of imbalance and linearity. Empirically we show the superiority of draGAN, but also highlight some of its shortcomings. \\
    All code is available on: \href{https://github.com/LeonGuertler/draGAN}{https://github.com/LeonGuertler/draGAN}
\end{abstract}


\end{frontmatter}

%\linenumbers

\section{Introduction}
    Numerous real-world business applications of Machine Learning are based on tabular data \cite{LESSMANN2015124, 1297040} which is often imbalanced \cite{LESSMANN2015124}.  
    This poses a significant problem to Machine Learning models, as they will oftentimes only fit the majority class, neglecting the minority class \cite{Ali2015ClassificationWC,doi:10.1142/S0218001409007326}. Alleviation of this problem can be achieved on both the data-level and the model-level. Whilst model-level approaches tend to focus on complex architectures or bespoke loss functions that pay more attention to the minority class \cite{5978225,RIVERA2016124}, data-level approaches are more general and focus on sampling methods \cite{Zhuoyuan}, data augmentation \cite{informatics7040052} and data generation \cite{RIVERA2016124,10.5555/1622407.1622416,NIPS2014_5ca3e9b1,ENGELMANN2021114582,last_et_al}, which makes them compatible with any classification model.
    SMOTE \cite{10.5555/1622407.1622416}, the ``de facto" standard for learning from imbalanced data, even though performing well in practice, is not able to effectively extend the training field \cite{10.1016/j.ins.2021.11.058}; generated samples lack diversity \cite{10.1016/j.ins.2021.11.058} and it does not take decision boundaries into account. Furthermore, any data points generated are solely based on the samples in the minority class, which means that the model misses out on valuable information about the data’s distribution.
    GAN based approaches, especially cWGAN \cite{ENGELMANN2021114582}, address the above issues but take substantially longer than SMOTE to converge. Moreover, one problem with using GANs to generate a batch of training samples for the classification model is that each data point is generated independently. Therefore, they might not be well distributed in the sample space and thus skew the classification model. However, the biggest shortcoming of both approaches is that they focus on generating synthetic data points that are similar to the real data, rather than data points that lead to the best classification performance.\\
    We propose a novel GAN based architecture, coined draGAN (Deep Reinforcement Augmented Generative Adversarial Network), that directly addresses the aforementioned problems. Namely, model collapse is solved by using one-dimensional convolutional layers in the Generator to map a single Gaussian noise vector to a whole training batch of samples, ensuring coherence; the objective of priming generated data points on the performance of the classification model is realized by using the Discriminator, in our work referred to as Critic, to estimate the performance (i.e. AUC) the classification model would achieve on the training data if it were trained on the generated data points.
    \\
    \\
    Our contribution can be summarized as follows:
    \begin{itemize}
        \item We create a new Generator architecture that generates full, coherent, training batches with every forward pass.
        \item We propose an alternative function of the adversarial, using it to estimate the training value of the generated data points, and therefore effectively extend its purpose into a domain beyond simply estimating the realisticity of generated data.
    \end{itemize}
    
    %All code is available on: https://github.com/LeonGuertler/draGAN
    
\section{Related Work}
    In SMOTE \cite{10.5555/1622407.1622416}, new datapoints ($x_{new}$) are generated by randomly sampling from the minority class ($x_i$) and creating a linear interpolation of its features with one of its nearest minority class neighbours ($x_j$). 
    
    \begin{equation}
        \label{eqn:SMOTE}
        x_{new} = x_i + \epsilon * (x_j - x_i); \text{where}\ \epsilon \sim U(0,1)
    \end{equation}

    \noindent This concept has be built upon by hundreds of similar algorithms and most of them were recently benchmarked by \cite{KOVACS2019105662}, where the authors point out that even though overall polynom-fit-SMOTE \cite{4670021} is the best one, the performance gain over SMOTE is marginal. Polynom-fit-SMOTE generates new data points along line segments between minority class samples that are connected via one of the four topological techniques ("bus", "star", "mesh" and "polynom"). This means that generated data points will be more scattered as compared to SMOTE. Though \cite{4670021} concludes that the "star" and "mesh" strategies perform best, \cite{KOVACS2019105662} finds that all four strategies perform similarly. In this work, we will utilize the "star" strategy.

    Similarly to SMOTE, MixUp \cite{zhang2018mixup} proposes a linear interpolation of features of two data points, however, these data points don't need to be from the same class or cluster, and hence the labels are also interpolated (\ref{eqn:mixup_x} and \ref{eqn:mixup_y} respectively). 
    
    \begin{equation}
        \label{eqn:mixup_x}
        x_{new} = \lambda x_i + (1-\lambda)x_j
    \end{equation}
    \begin{equation}
        \label{eqn:mixup_y}
        y_{new} = \lambda y_i + (1-\lambda)y_j
    \end{equation}
    
    Where $\lambda \sim \text{Beta}(\alpha, \alpha)$, for $\alpha \in (0, \infty)$.\\
    However, in the context of binary classification with out of the box models, the label has to be in a discrete format, which we enforce by rounding.
    \\
    Because of the nature of these algorithms, they are not able to effectively extend the training field \cite{10.1016/j.ins.2021.11.058}, generated samples can lack diversity \cite{10.1016/j.ins.2021.11.058} and decision boundaries are not taken into consideration. Moreover, they do not attempt to accurately approximate the probability distribution of the positive samples, which is an issue that GANs address.\\
    \\

    Since its inception, GAN \cite{NIPS2014_5ca3e9b1} has shown an astonishing capability of generating realistic samples for different, often high dimensional, data types \cite{8512865,8461714,8578241,DBLP:journals/corr/abs-2107-11098}. This popularity has resulted in the further development and application of the technique for other problems. 
    Recently, \cite{FIORE2019448} showed that GANs are a feasible solution to the imbalanced learning problem. However, they commonly suffer from unstable training \cite{DBLP:journals/corr/Goodfellow17}, have no quantitative way of indicating convergence and, as applied in \cite{FIORE2019448}, do not make use of all the available data, as the network is trained solely on the minority class.\\
    
    The instability of training and the absence of a convergence mechanism has successfully been addressed by a number of recent papers that utilize the Wasserstein-1 function as a loss for the discriminator, referred to as a critic in their work \cite{https://doi.org/10.48550/arxiv.1701.07875}. The discriminator has to lie within the space of 1-Lipschitz functions, which was enforced by clipping of weights. This both stabilizes the training process by providing more meaningful gradients and, since the Wasserstein-1 function measures the “distance”, shows whether the model converges.
    The method has been extended by \cite{NIPS2017_892c3b1c} where the 1-Lipschitz constraint was enforced through gradient clipping, which further improves the quality of the gradients, especially when the initial solution is far from the target solution. However, it should be pointed out that in the context of using a Generative model to improve the classification accuracy of a secondary model, the Wasserstein-1 function does not indicate the convergence towards the actual goal of the model, namely, improving the classification performance of the secondary model, but rather indicates how similar the generated samples are to the original data.
    
    The vanilla GAN model has no mechanism for generating samples with specific labels, except training the model exclusively on these labels. The idea of generating label specific data with GANs was first introduced by \cite{DBLP:journals/corr/MirzaO14} where the Generator and Discriminator accept a one-hot encoded label vector in addition to the usual inputs (\ref{eqn:cGAN}). This architecture, coined cGAN, was successfully used to generate MNIST digits for given labels.

    \begin{equation}
        \label{eqn:cGAN}
        \min_{G}\max_{D}\mathbb{E}_{x\sim p_{\text{data}}(x)}[\log{D(x|y)}] +  \mathbb{E}_{z\sim p_{\text{z}}(z)}[1 - \log{D(G(z|y))}]
    \end{equation}
    \\
    \\
    
    This method has been improved by an architecture coined AC-GAN \cite{https://doi.org/10.48550/arxiv.1610.09585} where, rather than providing two separate inputs, the Gaussian noise vector is augmented with a one-hot encoded label vector before being fed to the Generator. The Discriminator has two separate final layers, where one, the auxiliary head, predicts the label of the provided sample, and the other, the discriminatory head, whether the sample was generated. Whilst the Generator tries to maximize $L_C - L_S$, the Discriminator tries to maximize $L_C + L_S$.
    
    \begin{equation}
        \label{eqn:AC-GAN_Ls}
        L_S = \mathbb{E}[\log{P(S = \text{real} | X_{\text{real}})}] + \mathbb{E}[\log{P(S = \text{fake} | X_{\text{fake}})}]
    \end{equation}
    
    \begin{equation}
        \label{eqn:AC-GAN_Lc}
        L_C = \mathbb{E}[\log{P(C = c | X_{\text{real}})}] + \mathbb{E}[\log{(C = c | X_{\text{fake}})}]
    \end{equation}
    \\
    \\
    
    cWGAN as used in \cite{ENGELMANN2021114582} presents a combination of WGAN-GP \cite{NIPS2017_892c3b1c} and an AC-GAN \cite{https://doi.org/10.48550/arxiv.1610.09585}, by incorporating the AC loss. However, as they point out in their paper, their implementation deviates from the classic AC-GAN by using two separate networks for the discriminator and the auxiliary classifier, rather than merely two different heads. The architecture, which is optimized by (\ref{eqn:cWGAN}), was purpose-built for oversampling tabular data for imbalanced learning and, to the best of our knowledge, is currently the state-of-the-art GAN based model for the aforementioned problem.
    \begin{equation}
        \label{eqn:Wasserstein_loss}
        \text{Wasserstein-loss} = \mathbb{E}_{x\sim p_{\text{data}}(x)}[D(X)] - \mathbb{E}_{z\sim p_{z}}[D(G(z))]
    \end{equation}
    
    \begin{equation}
        \label{eqn:gradient_penalty}
        \text{Gradient-penalty} = \lambda_{GP}\mathbb{E}_{\hat{x}\sim p_{\hat{x}}}[(\|\nabla_{\hat{x}}D(\hat{x})\|_2-1)^2]
    \end{equation}

    \begin{equation}
        \label{eqn:AC loss}
        \text{AC-loss} = \lambda_{\text{AC}}\mathbb{E}_{z\sim p_z}[\text{BCE}(\text{AC}(G(z))))]
    \end{equation}
    
    \begin{equation}
        \label{eqn:cWGAN}
        \min_{G}\max_{D} (\text{Wasserstein-loss} - \text{Gradient-penalty} + \text{AC-loss})
    \end{equation}
    \\
    \\

    One key issue that is neither addressed by cWGAN \cite{ENGELMANN2021114582} nor SMOTE \cite{10.5555/1622407.1622416} is that for the purpose of building a high-performing classification model on an imbalanced dataset, the most important metric should be the model's performance. Thus, we propose a GAN based architecture that incorporates a deep reinforcement learning agent to estimate the quality, in regards to improved classification performance, of each generated data point, called draGAN.\\
    \par
    Before diving into the details of our proposed method (draGAN), we want to prevent any potential confusion by pointing out that this is a novel work, and in no way related to the DRAGAN algorithm developed by \cite{DRAGAN_trash}.

\section{Methodology}
    The purpose of supervised learning algorithms \footnote{Both the nomenclature and the original set-up of the first few equations is based on the content of \cite{DL_book}} is to predict a label y given an input x, for any pair that is jointly sampled from the distribution $p(x,y)$. This can be achieved by sampling a finite number of pairs $(x_{\tilde{\alpha}},y_{\tilde{\alpha}})_{\tilde{\alpha}\in \mathcal{A}}$ and minimizing
 
    \begin{equation}
        \label{eqn:general_loss_form}
        \mathcal{L}_{\mathcal{A}}(\theta) \equiv \sum_{\tilde{\alpha}\in \mathcal{A}} {\mathcal{L}(z(x_{\tilde{\alpha}};\theta),y_{\tilde{\alpha}})}
    \end{equation}
   
    where $\tilde{\alpha}\in \mathcal{A}$ denotes inputs from the training set and $\mathcal{L}$ a loss function. The quality of the fitted parameters $\theta$ is commonly estimated on a number of test inputs, denoted as $\dot{\beta}\in \mathcal{B}$. The difference of the test loss and the training loss is referred to as the Generalization-error        (\ref{eqn:generalization_error})
   
    \begin{equation}
       \label{eqn:generalization_error}
       \mathcal{E} = \mathcal{L}_{\mathcal{B}} - \mathcal{L}_{\mathcal{A}}
    \end{equation}
    \\
    In imbalanced classification problems, one class, referred to as the majority class, will be over-represented as compared to the other, referred to as minority class. This means that any equally weighted measures of success, such as accuracy, will be miss-leading. Thus, new measures that pay more attention to the minority class have been introduce \cite{https://doi.org/10.1002/asi.4630300621,inproceedings}. Inversely to the loss function, these measures indicate optimal performance at 1 and poor performance at 0.\\
   
   Unfortunately, since these performance indicators are discrete, they can not directly be used for training. Furthermore, any equally weighted metric used for training, such as the "de-facto" Negative Log-Likelihood Loss (\ref{eqn:NLL-Loss}), will not guarantee a strictly inversely correlated performance on the discrete metrics, which will ultimately be used to validated the models success.

    \begin{equation}
       \label{eqn:NLL-Loss}
       \mathcal{L}_{\mathcal{A}}(\theta) \equiv -\frac{1}{|\mathcal{A}|} \sum_{\tilde{\alpha}\in \mathcal{A}} { y_{\tilde{\alpha}} * log(z(x_{\tilde{\alpha}};\theta)) + (1-y_{\tilde{\alpha}}) * log(1-z(x_{\tilde{\alpha}};\theta)) }
    \end{equation}
   
    where $|\mathcal{A}|$ denote the cardinality (size) of the training set.\\
   
    We justify this claim by computing the trivial solution with the lowest test loss and the corresponding $\text{F}_1\text{-Score}$.\\

    We denote the weights corresponding to the proposed trivial solution as $\theta_{\text{trivial}}$. A trivial solution will have the best loss when a constant $\epsilon$ minimizes (\ref{eqn:nllloss_trivial}), for $\epsilon \in (0,1)$.

    \begin{align}
        \mathcal{L}_{\mathcal{B}}(\theta_{\text{trivial}}) \notag \\
        & \equiv -\frac{1}{|\mathcal{B}|} \sum_{\dot{\beta}\in \mathcal{B}} { y_{\dot{\beta}} * log(z(x_{\dot{\beta}};\theta_{\text{trivial}})) + (1-y_{\dot{\beta}}) * log(1-z(x_{\dot{\beta}};\theta_{\text{trivial}})) } \notag \\
        & = -\frac{1}{|\mathcal{B}|} \sum_{\dot{\beta}\in \mathcal{B}} { y_{\dot{\beta}} * log(\epsilon) + (1-y_{\dot{\beta}}) * log(1-\epsilon) } \label{eqn:nllloss_trivial}
    \end{align}
    \\
    $\epsilon$ can easily be determined by dividing (\ref{eqn:nllloss_trivial}) into a majority set ($\mathcal{B}_{\text{\tiny majority}} \subset \mathcal{B}$) and a minority set ($\mathcal{B}_{\text{\tiny minority}} \subset \mathcal{B}$), where $\mathcal{B}_{\text{\tiny minority}} = (\mathcal{B}_{\text{\tiny majority}})^{\complement}$ and hence $|\mathcal{B}_{\text{\tiny minority}}| + |\mathcal{B}_{\text{\tiny majority}}| = |\mathcal{B}|$.
    
    \begin{equation}
         \mathcal{L}_{\mathcal{B}}(\theta_{\text{trivial}}) = - \frac{1} {|\mathcal{B}|} * (|\mathcal{B}_{\text{\tiny minority}}|*log(\epsilon) + |\mathcal{B}_{\text{\tiny majority}}|*log(1 - \epsilon))
    \end{equation}
    
    \begin{align}
        \frac{d \mathcal{L}_{\mathcal{B}}(\theta_{\text{trivial}})} {d \epsilon} &= - \frac{1}{|\mathcal{B}|} * (\frac{|\mathcal{B}_{\text{\tiny minority}}|} {\epsilon} - \frac{|\mathcal{B}_{\text{\tiny majority}}|} {1-\epsilon})
    \end{align}
    \begin{equation}
        \therefore \text{Minimum at } \epsilon = \frac{|\mathcal{B}_{\text{\tiny minority}}|} {|\mathcal{B}|}
    \end{equation}\\

    The theoretical continuous confusion matrix can be expressed as.
    
    \begin{tabular}{@{}cc|cc@{}}
        \multicolumn{1}{c}{} &\multicolumn{1}{c}{} &\multicolumn{2}{c}{Predicted} \\ 
        \multicolumn{1}{c}{} & 
        \multicolumn{1}{c|}{} & 
        \multicolumn{1}{c}{Yes} & 
        \multicolumn{1}{c}{No} \\ 
        \cline{2-4}
        \multirow[c]{2}{*}{\rotatebox[origin=tr]{90}{Actual}}
        & Yes  & $0.5*\frac{|\mathcal{B}_{\text{\tiny minority}}|}{|\mathcal{B}|}* |\mathcal{B}|$ & $0.5*(1-\frac{|\mathcal{B}_{\text{\tiny minority}}|}{|\mathcal{B}|})* |\mathcal{B}|$   \\[1.5ex]
        & No  & $0.5*\frac{|\mathcal{B}_{\text{\tiny minority}}|}{|\mathcal{B}|}* |\mathcal{B}|$  & $0.5*(1-\frac{|\mathcal{B}_{\text{\tiny minority}}|}{|\mathcal{B}|})* |\mathcal{B}|$ \\ 
        \cline{2-4}
    \end{tabular}\\
    
\begin{figure}[h]
        \centering
        \includegraphics[width=0.75\textwidth]{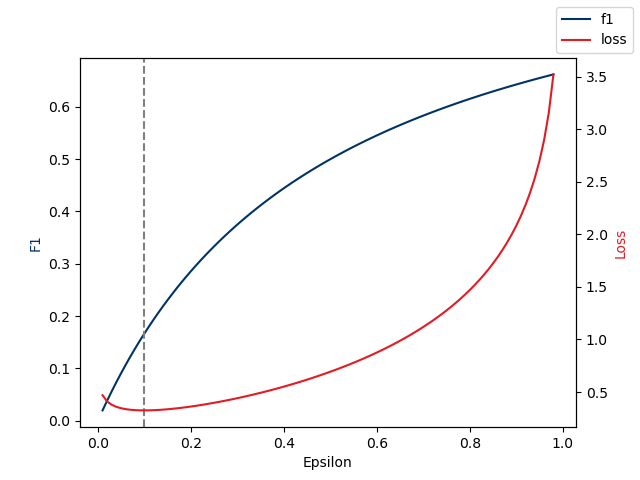}
        \caption{Loss vs $\text{F}_1\text{-Score}$}
        \label{fig:loss_vs_f1}
    \end{figure}

    Therefore, the $F_1\text{-Score}$ can be evaluated as
    
    \begin{align}
        \text{Precision}_{\mathcal{B}} &= \frac {0.5*\frac{|\mathcal{B}_{\text{\tiny minority}}|}{|\mathcal{B}|}* |\mathcal{B}|} {0.5*\frac{|\mathcal{B}_{\text{\tiny minority}}|}{|\mathcal{B}|}* |\mathcal{B}| + 0.5*\frac{|\mathcal{B}_{\text{\tiny minority}}|}{|\mathcal{B}|}* |\mathcal{B}|} = 0.5
    \end{align}
    
    \begin{align}
        \text{Recall}_{\mathcal{B}} &= \frac {0.5*\frac{|\mathcal{B}_{\text{\tiny minority}}|}{|\mathcal{B}|} * |\mathcal{B}|} {0.5*\frac{|\mathcal{B}_{\text{\tiny minority}}|}{|\mathcal{B}|}* |\mathcal{B}| + 0.5*(1-\frac{|\mathcal{B}_{\text{\tiny minority}}|}{|\mathcal{B}|})* |\mathcal{B}|} = \frac{|\mathcal{B}_{\text{\tiny minority}}|}{|\mathcal{B}|}
    \end{align}
    
    \begin{align}
        F_{1}\text{-Score} &= \frac{2*0.5*\frac {|\mathcal{B}_{\text{\tiny minority}}|} {|\mathcal{B}|}} {0.5 + \frac {|\mathcal{B}_{\text{\tiny minority}}|} {|\mathcal{B}|}} \label{eqn:f1_trivial} = \frac {|\mathcal{B}_{\text{\tiny minority}}|} {0.5*|\mathcal{B}|+|\mathcal{B}_{\text{\tiny minority}}|} = \frac{\epsilon}{0.5+\epsilon}
    \end{align}
    
    where (\ref{eqn:f1_trivial}) is a strictly increasing function with respect to the cardinality of $\mathcal{B}_{\text{\tiny minority}}$. This means that the higher the imbalance ratio, the lower the $F_{1}\text{-Score}$ of the trivial solution will be, which makes the function strictly correlated to the loss of the trivial solution (\ref{eqn:nllloss_trivial}) for the interval $(\frac{|\mathcal{B}_{\text{\tiny minority}}|}{|\mathcal{B}|}, 1)$. We illustrate this by plotting the $F_{1}\text{-Score}$ and the Loss as a function of epsilon (Figure \ref{fig:loss_vs_f1}), for an arbitrary toy dataset with 10\% minority samples and 90\% majority samples.\\
    
    Or, to express the results more clearly, as the number of minority samples approaches 0, the $F_{1}\text{-Score}$ of the trivial solution approaches 0 (\ref{eqn:f1_limit}), which compares rather poorly to the 0.5 achieved by random guessing.
    \begin{equation}
    \label{eqn:f1_limit}
        \lim_{|\mathcal{B}_{\text{\tiny minority}}|\to0}\; \frac {|\mathcal{B}_{\text{\tiny minority}}|} {0.5*|\mathcal{B}|+|\mathcal{B}_{\text{\tiny minority}}|} = 0
    \end{equation}\\
    
    The "de-facto" solution to this problem is to generate new data points in the vicinity of other data points with the same class. This will alleviate the problem to an extend, since the best trivial solution will be equal to random guessing ($\frac{|\mathcal{B}_{\text{\tiny minority}}|}{|\mathcal{B|}} = 0.5$). However, it does not guarantee that the loss (both test and train) is strictly inversely correlated to our discrete measure of success. Hence, we would like to generate new data ($ \breve \gamma \in \mathcal{Y}$) such that minimizing $\mathcal{L}_{\mathcal{A}\cup\mathcal{Y}}$ will maximize $F_{1}\text{-Score}_{\mathcal{A}}$ and, if (\ref{eqn:generalization_error}) is low, $F_{1}\text{-Score}_{\mathcal{B}}$.\\
    \\
    
    In other words, we want to generate $\mathcal{Y}$ such that $\mathcal{L}_{\mathcal{A}\cup\mathcal{Y}}$ is inversely correlated to $F_{1}\text{-Score}_{\mathcal{A}}$. We will refer to any discrepancies of the $\mathcal{L}_{\mathcal{A}}$ and $F_{1}\text{-Score}_{\mathcal{A}}$ as the Performance-Error (\ref{eqn:performance_error}). Furthermore, we expand the Generalization error (\ref{eqn:generalization_error}) to include the $F_{1}\text{-Score}$ (\ref{eqn:generalization_error_f1}) (or any other discrete measure of success).

    \begin{equation}
        \label{eqn:performance_error}
        \text{Performance-Error}_{\mathcal{A}} = 1 + \frac{\text{Cov}(F_{1}\text{-Score}_{\mathcal{A}}, \mathcal{L}_{\mathcal{A}})} {\sqrt{ \text{Var}(F_{1}\text{-Score}_{\mathcal{A}}) \text{Var}(\mathcal{L}_{\mathcal{A}}) }} 
    \end{equation}
    where $\mathcal{A}$ is interchangeable with $\mathcal{B}$.
    
    \begin{equation}
       \label{eqn:generalization_error_f1}
       \mathcal{E}_{F_{1}\text{-Score}} = F_{1}\text{-Score}_{\mathcal{B}} - F_{1}\text{-Score}_{\mathcal{A}}
    \end{equation}

    Thus, the overall objective is to minimize (\ref{eqn:both_errors}). In our approach, we simplify this by designing the algorithm to optimize for a high $F_{1}\text{-Score}$. Hence, reducing (\ref{eqn:both_errors}) to (\ref{eqn:generalization_error_f1}).

    \begin{equation}
        \label{eqn:both_errors}
        \text{Total-Error} = \text{Performance-Error}_{\mathcal{A}} + \mathcal{E}
    \end{equation}
    
    However, because of the discrete nature of the $F_{1}\text{-Score}$ it can not directly be used as a loss function for a data generation algorithm.
    
    We solved this problem by training a GAN, using the Discriminator to estimate the $F_{1}\text{-Score}_{\mathcal{A}}(\theta_{\mathcal{Y}})$ (where $\theta_{\mathcal{Y}}$ are the parameters of a classification model that was trained on $\mathcal{Y}$), and the Generator to map random noise to $\mathcal{Y}$.\\
    
    More formally, the Generator maps a random noise vector, denoted as $\kappa$, to a batch of training samples, denoted as $\mathcal{Y}$. We will denote a set of batches of generated data as $\mathbfcal{Y}$ which means $\breve \gamma \in \mathcal{Y} \in \mathbfcal{Y}$. Hence, the loss functions for the Discriminator (in our work referred to as Critic) and Generator, are (\ref{eqn:draGAN_discriminator_loss}) and (\ref{eqn:draGAN_generator_loss}) respectively.
    
    \begin{align}
        \mathcal{L}_{\mathcal{A}}(\theta_{\text{\tiny dis}}) \label{eqn:draGAN_discriminator_loss} \\
        & \equiv \sum_{\kappa \in \mathcal{K}} \mathcal{L}(z_{\text{\tiny dis}}(z_{\text{\tiny gen}}(\kappa; \theta_{\text{\tiny gen}}); \theta_{\text{\tiny dis}}), F_{1}\text{-Score}(y_{\mathcal{A}}, \theta_{z_{\text{\tiny gen}}(\kappa; \theta_{\text{\tiny gen}})})) \notag\\
        & \equiv \sum_{\mathcal{Y} \in \mathbfcal{Y}} \mathcal{L} (z_{\text{\tiny dis}}(\mathcal{Y}; \theta_{\text{\tiny dis}}), F_{1}\text{-Score}(y_{\mathcal{A}}, \theta_{\mathcal{Y}})) \notag
    \end{align}

    \begin{align}
        \mathcal{L}_{\mathcal{A}}(\theta_{\text{\tiny gen}}) \label{eqn:draGAN_generator_loss} \\
        & \equiv \sum_{\kappa \in \mathcal{K}} \mathcal{L}(z_{\text{\tiny gen}}(\kappa; \theta_{\text{\tiny gen}}), z_{\text{\tiny dis}}(z_{\text{\tiny gen}}(\kappa; \theta_{\text{\tiny gen}}); \theta_{\text{\tiny dis}})) \notag\\
        & \equiv \sum_{\mathcal{Y} \in \mathbfcal{Y}} \mathcal{L}(\mathcal{Y}, z_{\text{\tiny dis}}(\mathcal{Y}; \theta_{\text{\tiny dis}})) \notag
    \end{align}\\
    \par
    
    In the following three subsections we offer a more intuitive understanding of draGAN.

    \subsection*{Model Overview}
   Similar to the vanilla GAN architecture, the Generator, a Neural Network, creates new data points by accepting a Gaussian noise vector as input. In our case, the data points generated correspond to a batch of training samples. This generated data is then used to train the classification model (i.e. a Logistic Regression model). Said model is subsequently evaluated on the actual training data. The metric for evaluation can be chosen liberally, which we demonstrate by using a discrete AUC score. The achieved performance is mapped to the generated training data used and then utilized as an X,y pair to train the Critic. In other words, the Critic will, over time, learn how valuable, from a classification performance perspective, each generated data point will be, which allows it to provide a meaningful loss to the Generator. This comes with the fortunate side product that one can train the model on discrete metrics, which the Critic will turn into a continuous loss for the Generator. 
    
    This architecture directly addresses the aforementioned problem of not taking the classification value of the generated data points into consideration. However, as a vigilant reader will have noticed, draGAN does not optimize either network to generate realistic data points. Even though this might intuitively seem like a problem, as the network could conceivably generate data that skews the classification model into an unrealistic direction, we empirically show that it results in better classification performances of the model, which is the overall objective of this architecture.
    
    \begin{figure}[h]
      \centering
      \includegraphics[width=12cm]{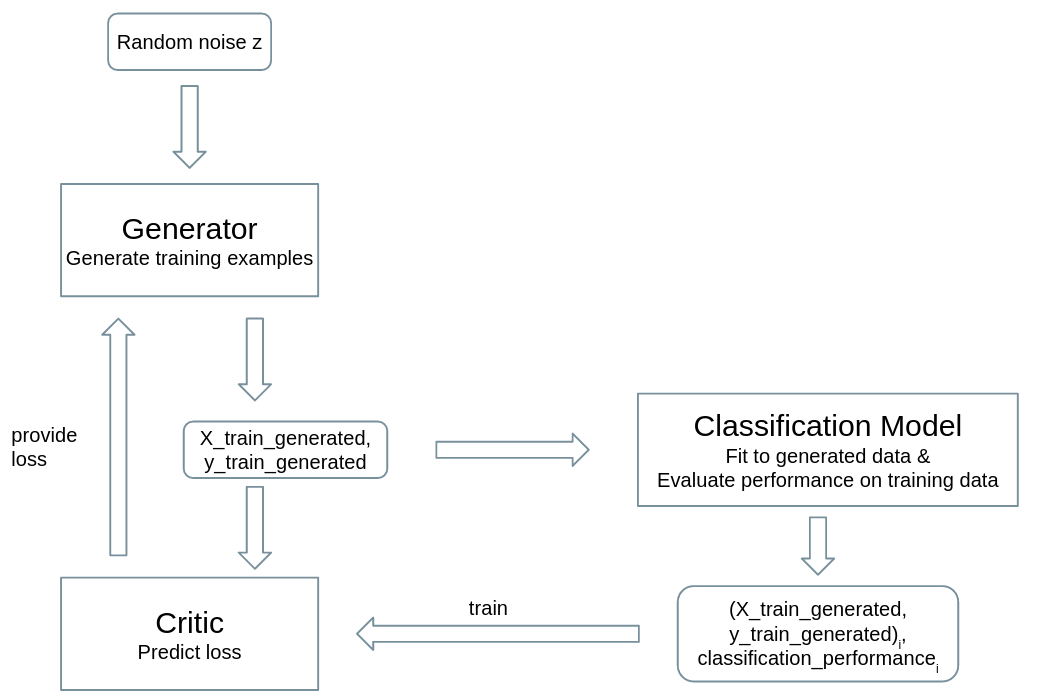}
      \caption{draGAN Architecture.}
    \end{figure}

\subsection{Generator}
    In Vanilla GAN, the Generator is used to map a random Gaussian noise vector to a sample that is hard to distinguish from the real data. However, since the network itself generates a single data point at a time, it oftentimes results in the Generator learning to map any random input to one, highly valued, data point \cite{DBLP:journals/corr/MetzPPS16}. When using a batch of generated samples to train the classification model, this poses an obvious problem. 
   
    Furthermore, even if the Generator does not collapse during training, the generated batch of data points might not be coherent. This means that the data points might be different from one another, and individually look realistic, but are not well distributed in the sample space. Therefore, a classification model trained on these generated data points will likely not perform well. 
    The Generator used in draGAN solves this issue by using a 1-dimensional convolutional layer to generate batches of samples at every forward pass. It follows that the loss will not be provided for any one single sample, but for the whole batch. This intuitively makes sense as one will ultimately use a batch of generated samples to train the classifier and not individual samples. 
    
\subsection{Adversarial}
    In vanilla GAN the adversarial, referred to as Discriminator, learns to classify whether a provided sample originated from the real data or was generated by the Generator. This process was further improved by the Wasserstein GAN where the adversarial, referred to as Critic, estimates the Wasserstein-1 function for generated values. I.e. how much “earth” would have to be moved from the distribution curve of the generated sample, to convert it to a real sample. Both of these approaches work well for generating realistic data points. However, that is not the objective when using GAN based approaches to improve classification performance, as realistic samples do not necessarily correspond to high classification performance.
    
    The adversarial in draGAN, referred to as Critic, addresses this problem by learning how to estimate the score, in our case AUC, that the classification model would achieve if it were trained on the generated data and assessed on the original training data. More concretely, the Critic learns to estimate the value of each generated batch with respect to the classifier's performance on a chosen metric. Therefore, throughout training, samples generated by the Generator will be increasingly helpful to the classifier in learning the distribution of the real training data, ultimately resulting in a better performance. 
    Furthermore, since the Critic will always provide a continuous loss to the Generator, even when trained on discrete values, draGAN can be trained to optimize the classifier's performance on any metric, including AUC. 
    This means that a classifier such as Logistic Regression can indirectly be optimized for any (discrete) metric.
    It is worth pointing out that similar to the Wasserstein GAN, draGAN has a mechanism for indicating convergence. Whilst the former uses the Wasserstein-1 function to measure the “distance” between generated and real samples, which gives a clear indication of whether the model is improving, the latter uses the Classification Models performance metric (i.e. AUC score) to indicate the same.
    
\subsection{Training Process}
    The training process of draGAN is typical for most GAN based architectures, where both the Generator and Adversarial are re-trained frequently and the Adversarial is used to provide some loss to the Generator. The key difference is that usually, the Adversarial learns to classify samples to be either generated or real \cite{NIPS2014_5ca3e9b1} or learns to estimate the similarity of a given sample to the real data \cite{https://doi.org/10.48550/arxiv.1701.07875}. In draGAN, however, the value of each generated sample is determined by training the classification model on the generated data and assessing its performance on the training data. Subsequently, these data points are used to train the Critic. 
    This process draws similarities to Reinforcement learning. For example, similarly to the memory recall method proposed in \cite{nature_dqn}, after each iteration, some of the Critic’s training data is saved for recalling in future training.
    \cite{https://doi.org/10.48550/arxiv.1701.07875} proposed to train the Critic until convergence in each iteration, because of the differentiability of Wasserstein-1, whilst \cite{NIPS2014_5ca3e9b1} determined a single training step to be the optimum. Because of the non-differentiability of the function that the Critic in draGAN attempts to estimate, we do not train the Critic until convergence but empirically estimated a good value to be in the range from 10 to 100 steps. This, however, is a hyperparameter and thus the optimum will be subject to the size of the network and the data used.

\section{Experimental Setup}
    To empirically evaluate the performance of draGAN, we benchmarked it on 94 imbalanced classification datasets (Table \ref{datasets-table}, in Appendix \ref{appendix:datasets}) against a plethora of state-of-the-art data generation algorithms from the SMOTE and GAN families. For these experiments we chose AUC to be the metric that draGAN is supposed to maximize.

\subsection{Preprocessing}
    \cite{KOVACS2019105662} is, to the best of our knowledge, the the most extensive empirical study of oversampling techniques. For ease of comparison between their work and ours, we utilize the exact same reprocessing and sampling pipeline.

\subsection{Datasets}
    $N_+$ denotes the number minority class samples and $N_-$ the number of majority class samples (note $N = N_+ + N_-$) and IR indicates the imbalance ratio (\ref{eqn:imbalance ratio}).  
    \begin{equation}
        \label{eqn:imbalance ratio}
        \text{IR} = \frac{N_-}{N_+}
    \end{equation}\\
    As proposed by \cite{ENGELMANN2021114582} we use the difference of AUC scores achieved by a Random Forest Classifier and a Logistic Regression as an estimator of the linearity of each dataset. If the difference is greater than 0.1, the dataset is considered non-Linear.\\
    An exhaustive table of all datasets used can in be found in Appendix \ref{appendix:datasets}

\subsection{Performance Measure}
    As already pointed out, the accuracy of a classification model on an imbalanced dataset gives a very poor indication of performance. Therefore, we use three alternative measures that are commonly utilized to evaluate performances for imbalanced classification tasks \cite{KOVACS2019105662,Raeder2012LearningFI,LOPEZ2013113,LOPEZ20141}. The notation is as follows, TP, TN, FP and FN correspond to true positive, true negative, false positive and false negative respectively.\\\\
    
    {\bfseries G-score}: which is sometimes referred to as Geometric Mean or G-Mean measures the balance between classification performances on both the majority and the minority classes.
    \begin{equation}
        \label{eqn:g-score}
        G = \sqrt{ \frac{\text{TP}}{\text{P}} *\frac{\text{TN}}{\text{N}} }
    \end{equation}
    Where P$=$TP$+$FN and N$=$TN$+$FP \\\\
    
    {\bfseries $\text{F}_1$-score}: (which is a specific form of the F-score) is the harmonic mean of precision (\ref{eqn:precision}) and recall (\ref{eqn:recall}).
    \begin{equation}
        \label{eqn:precision}
        \text{Precision} = \frac{\text{TP}}{\text{TP}+\text{FP}}
    \end{equation}
    
    \begin{equation}
        \label{eqn:recall}
        \text{Recall} = \frac{\text{TP}}{\text{TP}+\text{FN}}
    \end{equation}
    
    \begin{equation}
        \label{eqn:f-score}
        \text{F}_1 = 2*\frac {\text{Precision}*\text{Recall}} {\text{Precision}+\text{Recall}}
    \end{equation}\\\\
    
    {\bfseries AUC-Score}: which abbreviates \fb{Area} \fb{Under} the receiver operating characteristic \fb{Curve} quantifies the graph of sensitives plotted against corresponding false positive rates by calculating the area under the curve.
    
    \begin{equation}
        \label{eqn:TPR}
        \text{TPR} = \frac {\text{TP}} {\text{TP} + \text{FN}}
    \end{equation}
    
    \begin{equation}
        \label{eqn:FPR}
        \text{FPR} = \frac {\text{FP}} {\text{FP} + \text{TN}} 
    \end{equation}
    
    \begin{equation}
        \label{eqn:auc}
        \text{AUC} = \int_{0}^{1}{\text{TPR} (\text{FPR}^{-1}(x)) dx}
    \end{equation}

\subsection{Data-split}\label{sec:testing}
    Classification performance is evaluated by repeated stratified k-fold cross-validation with 5 splits and 3 repeats. 

\subsection{Hyperparameter tuning}
    For Hyperparameter tuning of draGAN, 15 datasets were randomly selected. Each datataset was stratified with 5 splits, omitting the test data and using a fraction of the training data as validation, so that we do not skew the empirical results. Parameters were sampled via the Tree-structured Parzen Estimator algorithm, as implemented in Optuna. A list of all hyperparameters can be found in the Appendix \ref{appendix:hyperparameters}.

\section{Results}
    In this section we will analyze the performances of our benchmarks as compared to draGAN using the AUC-score. Because of it's slow speed the cWGAN algorithm was only evaluated on a subset of the datasets. The cWGAN benchmarking and the results as evaluated on the other discrete measures of success can be found in the Appendix \ref{appendix:cwgan_results} and Appendix \ref{appendix:detailed_results} respectively.\\
    For ease of comparison, an "Average" row was added to Table \ref{table:results}.
    
    \begin{center}
    \centering
    \tiny
        \begin{longtable}{llccccc}
        \caption{Results}
            \label{table:results}\\
            Nr & Dataset                                  & Vanilla           & SMOTE             & Polynom\_fit\_SMOTE        & MixUp             & draGAN (ours) \\
            \hline
            0  & cm1                                      & 0.729             & 0.7292            & 0.7292                     & 0.7292            & \bf{0.738}             \\
            1  & german                                   & 0.5537            & 0.5537            & 0.5537                     & 0.5537            & \bf{0.661}             \\
            2  & hepatitis                                & 0.742             & 0.7428            & 0.7788                     & 0.7639            & \bf{0.783}             \\
            3  & hypothyroid                              & 0.8111            & 0.8445            & \bf{0.8483}                     & 0.826             & 0.8278            \\
            4  & kc1                                      & 0.7962            & 0.7962            & 0.7962                     & 0.7962            & \bf{0.7972}            \\
            5  & pc1                                      & 0.657             & 0.657             & 0.657                      & 0.657             & \bf{0.7122}            \\
            6  & satimage                                 & 0.7027            & 0.7057            & \bf{0.7153}                     & 0.7008            & 0.6902            \\
            7  & spectf                                   & \bf{0.8522}            & 0.8501            & 0.8509                     & \bf{0.8522}            & 0.8116            \\
            8  & abalone\_17\_vs\_7\_8\_9\_10             & 0.7512            & 0.8436            & 0.8486                     & 0.7626            & \bf{0.9304}            \\
            9  & abalone\_19\_vs\_10\_11\_12\_13          & 0.454             & 0.7277            & 0.7255                     & 0.5064            & \bf{0.8063}            \\
            10 & abalone\_20\_vs\_8\_9\_10                & 0.7548            & 0.9094            & 0.9167                     & 0.7703            & \bf{0.9747}            \\
            11 & abalone\_21\_vs\_8                       & 0.8957            & 0.9139            & 0.9157                     & 0.9014            & \bf{0.966}             \\
            12 & abalone\_3\_vs\_11                       & \bf{1}                 & \bf{1}                 & \bf{1}                          & \bf{1}                 & 0.9995            \\
            13 & abalone9\_18                             & 0.731             & 0.7801            & 0.7955                     & 0.7401            & \bf{0.938}             \\
            14 & car\_good                                & \bf{0.6223}            & 0.6127            & 0.6205                     & 0.6166            & 0.5922            \\
            15 & car\_vgood                               & 0.9308            & 0.9335            & \bf{0.9343}                     & 0.9296            & 0.9252            \\
            16 & cleveland\_0\_vs\_4                      & 0.6097            & 0.6536            & 0.6435                     & 0.6259            & \bf{0.6958}            \\
            17 & ecoli\_0\_1\_3\_7\_vs\_2\_6              & 0.9066            & 0.9189            & 0.9304                     & 0.8823            & \bf{0.9458}            \\
            18 & ecoli\_0\_1\_4\_6\_vs\_5                 & \bf{0.949}             & 0.9429            & 0.9298                     & 0.9449            & 0.9013            \\
            19 & ecoli\_0\_1\_4\_7\_vs\_2\_3\_5\_6        & 0.8461            & 0.8444            & 0.8351                     & \bf{0.8526}            & 0.8304            \\
            20 & ecoli\_0\_1\_4\_7\_vs\_5\_6              & 0.8914            & 0.887             & 0.875                      & \bf{0.8982}            & 0.8556            \\
            21 & ecoli\_0\_1\_vs\_2\_3\_5                 & \bf{0.9198}            & 0.9062            & 0.9143                     & 0.9143            & 0.8695            \\
            22 & ecoli\_0\_1\_vs\_5                       & \bf{0.9716}            & 0.9652            & 0.958                      & \bf{0.9716}            & 0.9121            \\
            23 & ecoli\_0\_2\_3\_4\_vs\_5                 & \bf{0.9408}            & 0.9232            & 0.9231                     & 0.9358            & 0.8617            \\
            24 & ecoli\_0\_2\_6\_7\_vs\_3\_5              & \bf{0.8645}            & 0.8434            & 0.8537                     & 0.861             & 0.7828            \\
            25 & ecoli\_0\_3\_4\_6\_vs\_5                 & \bf{0.9419}            & 0.9311            & 0.9203                     & 0.9311            & 0.8703            \\
            26 & ecoli\_0\_3\_4\_7\_vs\_5\_6              & \bf{0.8636}            & 0.8621            & 0.8532                     & 0.85              & 0.8538            \\
            27 & ecoli\_0\_3\_4\_vs\_5                    & \bf{0.9306}            & 0.9111            & 0.9056                     & 0.925             & 0.8361            \\
            28 & ecoli\_0\_4\_6\_vs\_5                    & \bf{0.9409}            & 0.9226            & 0.9148                     & 0.9345            & 0.8572            \\
            29 & ecoli\_0\_6\_7\_vs\_3\_5                 & \bf{0.8792}            & 0.8568            & 0.8622                     & 0.8767            & 0.8406            \\
            30 & ecoli\_0\_6\_7\_vs\_5                    & \bf{0.9425}            & 0.9342            & 0.9313                     & 0.9363            & 0.8937            \\
            31 & flaref                                   & 0.8854            & 0.9095            & \bf{0.9117}                     & 0.8952            & 0.9012            \\
            32 & glass\_0\_1\_4\_6\_vs\_2                 & 0.6351            & 0.6058            & 0.6351                     & 0.6314            & \bf{0.6645}            \\
            33 & glass\_0\_1\_5\_vs\_2                    & 0.6344            & 0.5685            & \bf{0.636}                      & 0.6185            & 0.6167            \\
            34 & glass\_0\_1\_6\_vs\_2                    & 0.6238            & 0.5868            & 0.631                      & 0.5978            & \bf{0.6452}            \\
            35 & glass\_0\_1\_6\_vs\_5                    & 0.86              & 0.8762            & 0.86                       & 0.8848            & \bf{0.94}              \\
            36 & glass\_0\_4\_vs\_5                       & \bf{0.9625}            & \bf{0.9625}            & \bf{0.9625}                     & \bf{0.9625}            & 0.8668            \\
            37 & glass\_0\_6\_vs\_5                       & 0.8379            & 0.8642            & 0.8379                     & 0.8262            & \bf{0.9711}            \\
            38 & glass\_0\_1\_2\_3\_vs\_4\_5\_6           & 0.9763            & 0.9758            & \bf{0.9769}                     & 0.9756            & 0.9714            \\
            39 & glass0                                   & 0.7306            & 0.7396            & 0.7345                     & 0.7286            & \bf{0.8339}            \\
            40 & glass1                                   & 0.5245            & \bf{0.5526}            & 0.525                      & 0.5368            & 0.5436            \\
            41 & glass6                                   & 0.971             & \bf{0.9713}            & 0.971                      & 0.9707            & 0.9569            \\
            42 & glass2                                   & 0.657             & 0.6228            & 0.6553                     & 0.6515            & \bf{0.7172}            \\
            43 & glass4                                   & 0.765             & 0.7847            & 0.765                      & 0.7719            & \bf{0.8308}            \\
            44 & glass5                                   & 0.8732            & 0.8878            & 0.8683                     & 0.8967            & \bf{0.9569}            \\
            45 & kddcup\_buffer\_overflow\_vs\_back       & \bf{1}                 & \bf{1}                 & \bf{1}                          & \bf{1}                 & \bf{1}                 \\
            46 & kddcup\_guess\_passwd\_vs\_satan         & \bf{0.9994}            & \bf{0.9994}            & \bf{0.9994}                     & \bf{0.9994}            & 0.9957            \\
            47 & kddcup\_land\_vs\_portsweep              & \bf{0.999}             & \bf{0.999}             & \bf{0.999}                      & \bf{0.999}             & 0.998             \\
            48 & kddcup\_land\_vs\_satan                  & 0.9994            & 0.9995            & 0.9995                     & 0.9994            & \bf{0.9997}            \\
            49 & kr\_vs\_k\_one\_vs\_fifteen              & \bf{1}                 & \bf{1}                 & \bf{1}                          & \bf{1}                 & \bf{1}                 \\
            50 & kr\_vs\_k\_three\_vs\_eleven1            & 0.9993            & 0.9996            & 0.9997                     & 0.9994            & \bf{0.9999}            \\
            51 & kr\_vs\_k\_zero\_one\_vs\_draw           & 0.9908            & 0.9881            & 0.9909                     & 0.9908            & \bf{0.9921}            \\
            52 & kr\_vs\_k\_zero\_vs\_eight               & 0.9066            & 0.9243            & \bf{0.9263}                     & 0.9067            & 0.9239            \\
            53 & kr\_vs\_k\_zero\_vs\_fifteen             & 0.999             & \bf{1}                 & \bf{1}                          & 0.9998            & 0.9995            \\
            54 & led7digit\_0\_2\_4\_5\_6\_7\_8\_9\_vs\_1 & 0.914             & 0.9145            & 0.9109                     & \bf{0.9162}            & 0.8916            \\
            55 & lymphography\_normal\_fibrosis           & 0.9324            & \bf{0.9929}            & \bf{0.9929}                     & 0.9369            & 0.8893            \\
            56 & page\_blocks\_1\_3\_vs\_4                & 0.9381            & 0.9389            & 0.9385                     & 0.9385            & \bf{0.95}              \\
            57 & poker\_8\_9\_vs\_51                      & 0.561             & 0.4873            & 0.5306                     & \bf{0.5675}            & 0.5196            \\
            58 & poker\_8\_9\_vs\_6                       & \bf{0.4495}            & 0.4421            & 0.4114                     & 0.4336            & 0.3993            \\
            59 & poker\_8\_vs\_6                          & 0.289             & 0.2863            & 0.2385                     & 0.268             & \bf{0.3229}            \\
            60 & poker\_9\_vs\_7                          & \bf{0.6427}            & 0.5755            & 0.5439                     & 0.6085            & 0.5607            \\
            61 & shuttle\_2\_vs\_5                        & 0.9928            & 0.9932            & 0.9927                     & 0.9916            & \bf{0.9966}            \\
            62 & shuttle\_6\_vs\_2\_3                     & \bf{1}                 & \bf{1}                 & \bf{1}                          & \bf{1}                 & 0.9977            \\
            63 & shuttle\_c0\_vs\_c4                      & 0.9879            & 0.9899            & 0.9891                     & 0.9881            & \bf{0.9999}            \\
            64 & shuttle\_c2\_vs\_c4                      & 0.975             & \bf{0.9806}            & 0.975                      & 0.9736            & 0.9737            \\
            65 & vowel0                                   & 0.9625            & 0.9633            & \bf{0.9641}                     & 0.963             & 0.9549            \\
            66 & winequality\_red\_3\_vs\_5               & 0.7944            & 0.792             & 0.7915                     & \bf{0.7947}            & 0.71              \\
            67 & winequality\_red\_4                      & 0.6089            & 0.6126            & 0.6162                     & 0.6096            & \bf{0.6764}            \\
            68 & winequality\_red\_8\_vs\_6               & 0.6357            & 0.6489            & 0.6552                     & 0.6381            & \bf{0.8228}            \\
            69 & winequality\_red\_8\_vs\_6\_7            & 0.6007            & 0.607             & 0.6138                     & 0.6013            & \bf{0.7653}            \\
            70 & winequality\_white\_3\_9\_vs\_5          & 0.3703            & 0.3474            & 0.3404                     & 0.4096            & \bf{0.6509}            \\
            71 & winequality\_white\_3\_vs\_7             & 0.6185            & 0.5724            & 0.6165                     & \bf{0.6185}            & 0.5269            \\
            72 & winequality\_white\_9\_vs\_4             & 0.6367            & 0.709             & 0.7114                     & 0.6256            & \bf{0.8502}            \\
            73 & yeast\_0\_2\_5\_6\_vs\_3\_7\_8\_9        & 0.8358            & 0.8208            & 0.8327                     & \bf{0.838}             & 0.7835            \\
            74 & yeast\_0\_2\_5\_7\_9\_vs\_3\_6\_8        & \bf{0.8877}            & 0.8848            & 0.8858                     & 0.8842            & 0.8327            \\
            75 & yeast\_0\_3\_5\_9\_vs\_7\_8              & 0.7468            & 0.7288            & \bf{0.751}                      & 0.7373            & 0.7122            \\
            76 & yeast\_0\_5\_6\_7\_9\_vs\_4              & 0.8307            & \bf{0.8392}            & 0.8309                     & 0.8312            & 0.8211            \\
            77 & yeast\_1\_2\_8\_9\_vs\_7                 & 0.7508            & 0.7813            & 0.7838                     & 0.7524            & \bf{0.79}              \\
            78 & yeast\_1\_4\_5\_8\_vs\_7                 & 0.6827            & 0.6784            & 0.6758                     & \bf{0.6881}            & 0.6343            \\
            79 & yeast\_1\_vs\_7                          & 0.8136            & \bf{0.8463}            & 0.8442                     & 0.8299            & 0.8429            \\
            80 & yeast\_2\_vs\_4                          & 0.9161            & \bf{0.9333}            & 0.9286                     & 0.9196            & 0.9312            \\
            81 & yeast\_2\_vs\_8                          & 0.844             & 0.8349            & 0.8423                     & \bf{0.8505}            & 0.7751            \\
            82 & yeast4                                   & 0.8709            & 0.8814            & 0.8811                     & 0.8745            & \bf{0.8909}            \\
            83 & yeast5                                   & 0.9861            & 0.9861            & 0.9864                     & 0.9861            & \bf{0.9874}            \\
            84 & yeast6                                   & 0.9249            & 0.9299            & \bf{0.9359}                     & 0.9279            & 0.927             \\
            85 & yeast1                                   & 0.7803            & 0.7808            & \bf{0.7834}                     & 0.7828            & 0.7749            \\
            86 & yeast3                                   & 0.9629            & 0.9653            & \bf{0.9654}                     & 0.9644            & 0.9336            \\
            87 & zoo\_3                                   & 0.7837            & \bf{0.8272}            & 0.8026                     & 0.7895            & 0.7695            \\
            88 & iris0                                    & \bf{1}                 & \bf{1}                 & \bf{1}                          & \bf{1}                 & \bf{1}                 \\
            89 & new\_thyroid1                            & 0.8857            & 0.8976            & 0.8992                     & 0.8915            & \bf{0.996}             \\
            90 & pima                                     & 0.7489            & 0.7621            & 0.774                      & 0.7551            & \bf{0.8088}            \\
            91 & segment0                                 & 0.8474            & 0.8611            & 0.8647                     & 0.8538            & \bf{0.964}             \\
            92 & vehicle0                                 & 0.8194            & 0.8231            & 0.8259                     & 0.8219            & \bf{0.9648}            \\
            93 & wisconsin                                & 0.995             & \bf{0.9951}            & 0.9948                     & 0.995             & 0.9939            \\
            \hline 
               &   \bf{Average}                                       & \bf{0.8195}    & \bf{0.8259}  & \bf{0.827}                & \bf{0.8207}   & \bf{0.8391}
        \end{longtable}
    \end{center}

    As can be observed in Table \ref{table:results}, draGAN achieved the highest AUC score most frequently (42 times) followed by Vanilla (25 times), polynom\_fit\_SMOTE (22 times), MixUp (19 times) and SMOTE (18 times). Beyond simple ranking, draGAN achieved the overall highest average AUC score and a more than 2.6 times higher performance gain over the baseline (Vanilla) than the current state-of-the-art (polynom\_fit\_SMOTE).
    
    \subsection{Correlation of results}
    To gain a better understanding of how similar certain algorithms perform, we calculated the Pearson Correlation Coefficient w.r.t. vanilla logistic regression (\ref{eqn:person_corr}) for each one. 
    \begin{equation}
        \label{eqn:person_corr}
        r_i = \frac{\sum_k(x_{i,k}-\bar x_i)(y_k-\bar y)}{\sqrt{\sum_k(x_{i,k}-\bar x_i)^2 \sum_k(y_k-\bar y)^2}}
    \end{equation}
    
    where $x_{i,k}$ denotes the performance achieved by algorithm i on dataset k, and $y_k$ the performance achieved by the vanilla logistic regression on dataset k.\\
    
    What is interesting is that the performances of SMOTE, polynom\_fit\_SMOTE and MixUp are all highly correlated to the performance of the vanilla Logistic Regression (0.97, 0.97 and 0.99 respectively), whilst draGAN performs more independently (0.86).

    \subsection{Highest Performance gains over Vanilla Logistic Regression}
    To determine whether it is worth running draGAN in practise, we compare the individual top 10 percentage point performance gains of each algorithm over vanilla Logistic Regression.

    \begin{table}[h]
        \centering
        \begin{tabular}{cccc}
             SMOTE & polynom\_fit\_SMOTE & MixUp & draGAN (ours)  \\
             \hline
             0.2737 & 0.2715 & 0.0524 & 0.3523 \\
             0.1546 & 0.1619 & 0.0393 & 0.2806 \\
             0.0924 & 0.0974 & 0.0248 & 0.2199 \\
             0.0723 & 0.0747 & 0.0235 & 0.2135 \\
             0.0605 & 0.0645 & 0.0219 & 0.207  \\
             0.0491 & 0.0605 & 0.0163 & 0.1871 \\
             0.0439 & 0.0372 & 0.0162 & 0.1792 \\
             0.0435 & 0.0368 & 0.0155 & 0.1646 \\
             0.0334 & 0.0338 & 0.0149 & 0.1454 \\
             0.0327 & 0.033  & 0.0123 & 0.1332 \\
             \hline
             \bf{0.0856} & \bf{0.0871} & \bf{0.0237} & \bf{0.2083}
        \end{tabular}
        \caption{Individual Top 10 percentage point gains over Vanilla Logistic Regression; the last line represents the averages of the above figures}
        \label{tab:pct_point_gain}
    \end{table}

    As can be observed in Table \ref{tab:pct_point_gain}, draGAN has both the single highest percentage point performance gain, and the highest average percentage point gain (out of the individual top 10 performances).\\
    As a vigilant reader will have realized, the only way such a substantial performance gain and yet only a comparatively marginal overall average performance gain is possible, is when the algorithm under-performs the baseline on other datasets. Further research is required to figure out what exactly causes the algorithm to over/under perform and potentially stabilize it. However, as of now, it seems advisable for practitioners to train both draGAN and a more stable SMOTE variant, before choosing the final sampling method to be used.\\
    
    As shown in the Appendix \ref{appendix:cwgan_results}, draGAN, converges much faster than other GAN based methods and has fewer tune-able parameters. Hence, the overall temporal and computational cost associated with adding the algorithm to ones toolkit, is marginal, whilst the potential performance gain can be substantial. \\
    
    \subsection{Classification Performance as a function of available datapoints}
\begin{figure}[h]
        \centering
        \includegraphics[width=\linewidth]{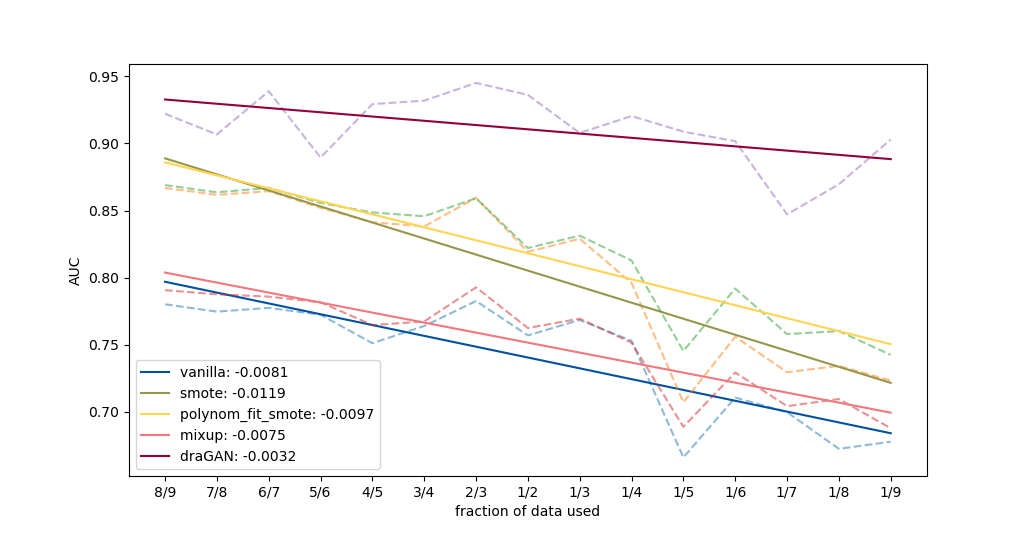}
        \caption{AUC score with varying data size}
        \label{fig:score_vs_samples}
    \end{figure}

    Lastly, we investigate the performance drop-off experienced by the various classifiers with respect to the number of samples available. Figure \ref{fig:score_vs_samples} visualizes the relationship of the classifiers performances with respect to the fraction of the dataset, abalone\_17\_vs\_7\_8\_9\_10, used. The numbers given in the legend are the slopes of the best fit line for each classifier. draGAN has, by far, the smallest absolute slope, which indicates that it experiences the smallest drop-off in performance when only very limited data is available. This is likely because it makes use of all available data, not just the minority class, and is able extrapolate.

\section{Conclusion}
    Even though we have successfully shown that our novel, task-oriented, data generation method, coined draGAN, is able to outperform the current state-of-the-art for generating new data points that improve the classification accuracy of a secondary model, we want to highlight the significant difference in approach to other GAN based models, rather than just the performance. To the best of our knowledge, this was the first time the adversarial was used to estimate the value of a generated data point in a domain beyond judging its realisticity. Since this approach is not limited to the aforementioned problem nor to any specific data type we hope to see its applications in different works in the future.

\bibliography{mybibfile}

\section*{Authors' contributions}
    \begin{itemize}
        \item \textit{Leon Guertler}: Ideation, conceptualization, implementation; Mathematical justifications; first draft 
        \item \textit{Andri Ashfahani}: Feedback \& suggestions
        \item \textit{Anh Tuan Luu}: Academic Supervision 
    \end{itemize}

\section*{Appendix}
    \subsection{Datasets}
    \label{appendix:datasets}
    \begin{center}
    \tiny
      \centering
      \begin{longtable}{llcccc}
        \caption{Datasets} 
        \label{datasets-table} \\
        \endfirsthead
        \endhead
        \toprule
        \bf ~ & \bf Dataset & \bf N & \bf \#Features & \bf IR & \bf non-Linear \\
        \midrule
            0  & cm1                                      & 498       & 23         & 9.16   & 0          \\
            1  & german                                   & 1000      & 29         & 2.33   & 0          \\
            2  & hepatitis                                & 155       & 19         & 3.84   & 0          \\
            3  & hypothyroid                              & 3163      & 25         & 19.95  & 1          \\
            4  & kc1                                      & 2109      & 21         & 5.47   & 0          \\
            5  & pc1                                      & 1109      & 21         & 13.40  & 0          \\
            6  & satimage                                 & 6435      & 36         & 9.28   & 1          \\
            7  & spectf                                   & 267       & 44         & 3.85   & 0          \\
            8  & abalone\_17\_vs\_7\_8\_9\_10             & 2338      & 8          & 39.31  & 0          \\
            9  & abalone\_19\_vs\_10\_11\_12\_13          & 1622      & 8          & 49.69  & 0          \\
            10 & abalone\_20\_vs\_8\_9\_10                & 1916      & 8          & 72.69  & 0          \\
            11 & abalone\_21\_vs\_8                       & 581       & 8          & 40.5   & 1          \\
            12 & abalone\_3\_vs\_11                       & 502       & 8          & 32.47  & 1          \\
            13 & abalone9\_18                             & 731       & 8          & 16.40  & 0          \\
            14 & car\_good                                & 1728      & 6          & 24.04  & 1          \\
            15 & car\_vgood                               & 1728      & 6          & 25.58  & 1          \\
            16 & cleveland\_0\_vs\_4                      & 177       & 23         & 12.62  & 0          \\
            17 & ecoli\_0\_1\_3\_7\_vs\_2\_6              & 281       & 7          & 39.14  & 0          \\
            18 & ecoli\_0\_1\_4\_6\_vs\_5                 & 280       & 6          & 13.0   & 0          \\
            19 & ecoli\_0\_1\_4\_7\_vs\_2\_3\_5\_6        & 336       & 7          & 10.59  & 0          \\
            20 & ecoli\_0\_1\_4\_7\_vs\_5\_6              & 332       & 6          & 12.28  & 0          \\
            21 & ecoli\_0\_1\_vs\_2\_3\_5                 & 244       & 7          & 9.17   & 0          \\
            22 & ecoli\_0\_1\_vs\_5                       & 240       & 6          & 11.0   & 0          \\
            23 & ecoli\_0\_2\_3\_4\_vs\_5                 & 202       & 7          & 9.1    & 0          \\
            24 & ecoli\_0\_2\_6\_7\_vs\_3\_5              & 224       & 7          & 9.18   & 0          \\
            25 & ecoli\_0\_3\_4\_6\_vs\_5                 & 205       & 7          & 9.25   & 0          \\
            26 & ecoli\_0\_3\_4\_7\_vs\_5\_6              & 257       & 7          & 9.28   & 0          \\
            27 & ecoli\_0\_3\_4\_vs\_5                    & 200       & 7          & 9.0    & 0          \\
            28 & ecoli\_0\_4\_6\_vs\_5                    & 203       & 6          & 9.15   & 0          \\
            29 & ecoli\_0\_6\_7\_vs\_3\_5                 & 222       & 7          & 9.09   & 0          \\
            30 & ecoli\_0\_6\_7\_vs\_5                    & 220       & 6          & 10.0   & 0          \\
            31 & flaref                                   & 1066      & 11         & 23.79  & 0          \\
            32 & glass\_0\_1\_4\_6\_vs\_2                 & 205       & 9          & 11.06  & 0          \\
            33 & glass\_0\_1\_5\_vs\_2                    & 172       & 9          & 9.12   & 0          \\
            34 & glass\_0\_1\_6\_vs\_2                    & 192       & 9          & 10.29  & 0          \\
            35 & glass\_0\_1\_6\_vs\_5                    & 184       & 9          & 19.44  & 1          \\
            36 & glass\_0\_4\_vs\_5                       & 92        & 9          & 9.22   & 1          \\
            37 & glass\_0\_6\_vs\_5                       & 108       & 9          & 11.0   & 1          \\
            38 & glass\_0\_1\_2\_3\_vs\_4\_5\_6           & 214       & 9          & 3.20   & 0          \\
            39 & glass0                                   & 214       & 9          & 2.06   & 1          \\
            40 & glass1                                   & 214       & 9          & 1.82   & 1          \\
            41 & glass6                                   & 214       & 9          & 6.38   & 0          \\
            42 & glass2                                   & 214       & 9          & 11.59  & 0          \\
            43 & glass4                                   & 214       & 9          & 15.46  & 1          \\
            44 & glass5                                   & 214       & 9          & 22.78  & 1          \\
            45 & kddcup\_buffer\_overflow\_vs\_back       & 2233      & 31         & 73.43  & 0          \\
            46 & kddcup\_guess\_passwd\_vs\_satan         & 1642      & 38         & 29.98  & 0          \\
            47 & kddcup\_land\_vs\_portsweep              & 1061      & 40         & 49.52  & 0          \\
            48 & kddcup\_land\_vs\_satan                  & 1610      & 30         & 75.67  & 0          \\
            49 & kr\_vs\_k\_one\_vs\_fifteen              & 2244      & 6          & 27.77  & 0          \\
            50 & kr\_vs\_k\_three\_vs\_eleven1            & 2935      & 6          & 35.23  & 0          \\
            51 & kr\_vs\_k\_zero\_one\_vs\_draw           & 2901      & 6          & 26.63  & 0          \\
            52 & kr\_vs\_k\_zero\_vs\_eight               & 1460      & 6          & 53.07  & 1          \\
            53 & kr\_vs\_k\_zero\_vs\_fifteen             & 2193      & 6          & 80.22  & 0          \\
            54 & led7digit\_0\_2\_4\_5\_6\_7\_8\_9\_vs\_1 & 443       & 7          & 10.97  & 0          \\
            55 & lymphography\_normal\_fibrosis           & 148       & 23         & 23.67  & 0          \\
            56 & page\_blocks\_1\_3\_vs\_4                & 472       & 10         & 15.86  & 0          \\
            57 & poker\_8\_9\_vs\_51                      & 2075      & 25         & 82.0   & 0          \\
            58 & poker\_8\_9\_vs\_6                       & 1485      & 25         & 58.4   & 0          \\
            59 & poker\_8\_vs\_6                          & 1477      & 25         & 85.88  & 0          \\
            60 & poker\_9\_vs\_7                          & 244       & 25         & 29.5   & 0          \\
            61 & shuttle\_2\_vs\_5                        & 3316      & 9          & 66.67  & 0          \\
            62 & shuttle\_6\_vs\_2\_3                     & 230       & 9          & 22.0   & 0          \\
            63 & shuttle\_c0\_vs\_c4                      & 1829      & 9          & 13.87  & 0          \\
            64 & shuttle\_c2\_vs\_c4                      & 129       & 9          & 20.5   & 0          \\
            65 & vowel0                                   & 988       & 13         & 9.978  & 0          \\
            66 & winequality\_red\_3\_vs\_5               & 691       & 11         & 68.1   & 0          \\
            67 & winequality\_red\_4                      & 1599      & 11         & 29.17  & 0          \\
            68 & winequality\_red\_8\_vs\_6               & 656       & 11         & 35.44  & 0          \\
            69 & winequality\_red\_8\_vs\_6\_7            & 855       & 11         & 46.5   & 0          \\
            70 & winequality\_white\_3\_9\_vs\_5          & 1482      & 11         & 58.28  & 0          \\
            71 & winequality\_white\_3\_vs\_7             & 900       & 11         & 44.0   & 0          \\
            72 & winequality\_white\_9\_vs\_4             & 168       & 11         & 32.6   & 1          \\
            73 & yeast\_0\_2\_5\_6\_vs\_3\_7\_8\_9        & 1004      & 10         & 9.14   & 1          \\
            74 & yeast\_0\_2\_5\_7\_9\_vs\_3\_6\_8        & 1004      & 10         & 9.14   & 1          \\
            75 & yeast\_0\_3\_5\_9\_vs\_7\_8              & 506       & 10         & 9.12   & 0          \\
            76 & yeast\_0\_5\_6\_7\_9\_vs\_4              & 528       & 10         & 9.35   & 1          \\
            77 & yeast\_1\_2\_8\_9\_vs\_7                 & 947       & 10         & 30.57  & 0          \\
            78 & yeast\_1\_4\_5\_8\_vs\_7                 & 693       & 10         & 22.1   & 0          \\
            79 & yeast\_1\_vs\_7                          & 459       & 7          & 14.3   & 0          \\
            80 & yeast\_2\_vs\_4                          & 514       & 8          & 9.08   & 1          \\
            81 & yeast\_2\_vs\_8                          & 482       & 10         & 23.1   & 0          \\
            82 & yeast4                                   & 1484      & 10         & 28.10  & 0          \\
            83 & yeast5                                   & 1484      & 10         & 32.73  & 1          \\
            84 & yeast6                                   & 1484      & 10         & 41.4   & 1          \\
            85 & yeast1                                   & 1484      & 10         & 2.46   & 0          \\
            86 & yeast3                                   & 1484      & 10         & 8.10   & 1          \\
            87 & zoo\_3                                   & 101       & 16         & 19.2   & 0          \\
            88 & iris0                                    & 150       & 4          & 2.0    & 0          \\
            89 & new\_thyroid1                            & 215       & 5          & 5.14   & 0          \\
            90 & pima                                     & 768       & 8          & 1.87   & 0          \\
            91 & segment0                                 & 2308      & 23         & 6.02   & 0          \\
            92 & vehicle0                                 & 846       & 18         & 3.25   & 0          \\
            93 & wisconsin                                & 683       & 9          & 1.86   & 0  
      \end{longtable}
    \end{center}

    \subsection{draGAN Hyperparameter tuning}
    \label{appendix:hyperparameters}
    
    Table \ref{table:hyperparams-dragan} shows the hyperparameters that performed best during the non-exhaustive Hyperparameter-tuning and were used for all experiments. All ambiguous hyperparameters are described in greater detail below the table.  
    
    \begin{center}
        \centering
        \tiny
        \begin{longtable}{lc}
            \caption{Hyperparameters (draGAN)} 
            \label{table:hyperparams-dragan}\\
                  
            \hline
            z\_size                         & 512 \\
            Generator Learning-Rate         & 0.000266 \\
            Generator Optimizer             & RMSprop \\
            Generator Activation            & Sigmoid \\
            Generator BatchNorm             & No \\
            Generator Dropout               & Yes \\
            
            Critic Learning-Rate            & 0.036284 \\
            Critic Nr. Epochs               & 2 \\
            Critic Optimizer                & Adam \\
            Critic Layer 1                  & 64 \\
            Critic Layer 2                  & 128 \\
            Critic Layer 3                  & 64 \\
            Critic Activation 1             & ReLU \\
            Critic Activation 2             & ReLU \\
            Critic Activation 3             & LeakyReLU \\
            Critic BatchNorm Layer 1        & Yes \\
            Critic BatchNorm Layer 2        & No \\ 
            Critic Dropout Layer 1          & No \\
            Critic Dropout Layer 2          & Yes \\
            
            Nr. Samples Generated factor    & 1.793469 \\
            Total Epochs                    & 1750 \\
            Batch Size                      & 16 \\
            Max Memory Factor               & 124 \\
            Early Stopping                  & 921 \\ 
            \hline
        \end{longtable}
    \end{center}
    
    \begin{itemize}
        \item \textit{Critic Nr. Epochs}: Similarly to the training process of the original GAN architecture, the Critic is re-trained multiple times for each training step of the Generator. 
        \item \textit{Nr. Samples Generated factor}: This is the factor used to determine how many samples draGAN should generated, based on the number of training samples provided.
        \item \textit{Max Memory Factor}: Similarly to the DQN architecture first proposed by \cite{nature_dqn}, in each iteration, we retain some of the Critic's training data for the next training iterations. The Max Memory Factor poses an upper limit to the number of samples in the models ``memory", which is enforced by deleting random elements thereof, rather than FIFO or FILO.
        \item \textit{Early Stopping}: In draGAN, early stopping is evaluated based on for how many epochs the discrete measure of success (i.e. AUC) has not improved.
    \end{itemize}

\begin{figure}[h!]
        \hfill
        \subfigure[Generator]{\includegraphics[scale=0.4]{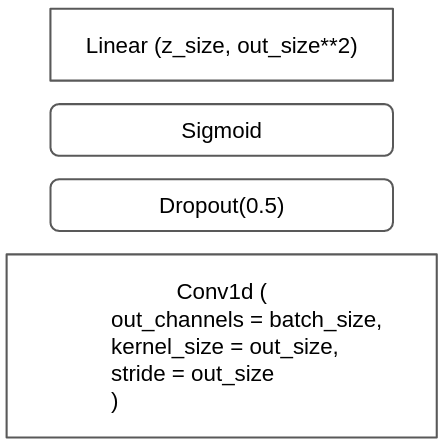}}
        \hfill
        \subfigure[Critic]{\includegraphics[scale=0.4]{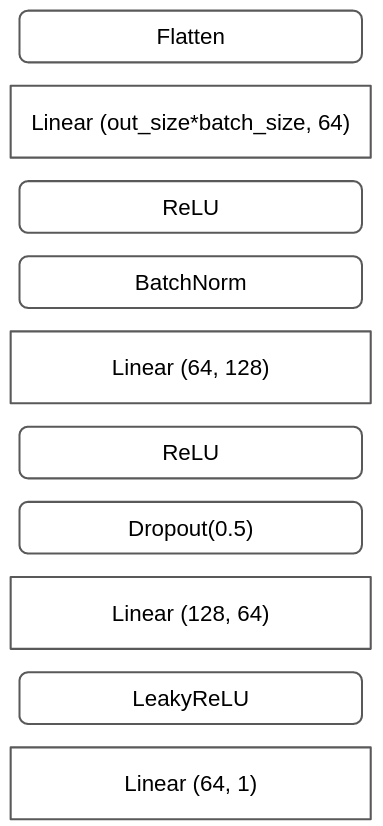}}
        \hfill
        \caption{draGAN Architecture}
        \label{fig:draGAN-architecture}
    \end{figure}

    Figure \ref{fig:draGAN-architecture} shows the exact architecture of the Generator and Critic in a diagrammatic form.

    \subsection{Experiment Hardware}
    \label{appendix:experiment_hardware}
    For reproduction of our results, and specifically because we report the time take by each algorithm, below is a list of the hardware used for all experiments.
    \begin{itemize}
        \item AMD Threadripper 3970x
        \item NVIDIA GeForce RTX 3090
        \item RAM DDR4 3600 (128gb)
        \item Ubuntu 20.04
    \end{itemize}

    \subsection{cWGAN results}
    \label{appendix:cwgan_results}
    As can be observed in Table \ref{table:cWGAN-time}, cWGAN, on average, takes more than 30-times as long to converge as draGAN. However, it is worthwhile pointing out that the library used, namely the one published by \cite{ENGELMANN2021114582}, runs on CPU whilst draGAN utilizes a GPU.\\
    
    \begin{center}
    \centering
    \tiny
        \begin{longtable}{llccccc}
        \caption{Average Time (in seconds)}
        \label{table:cWGAN-time}\\
        Nr & Dataset                                  & Vanilla           & SMOTE             & Polynom\_fit\_SMOTE        & MixUp             & draGAN (ours) \\
            \hline
            cm1                             & 0.0039         & 0.0057       & 0.006                      & 0.0057       & 127.4778      & 528.7398     \\
            german                          & 0.0045         & 0.0074       & 0.0063                     & 0.008        & 194.2204      & 1070.6063    \\
            hepatitis                       & 0.0046         & 0.0061       & 0.0052                     & 0.0045       & 77.2097       & 160.3711     \\
            hypothyroid                     & 0.0089         & 0.0178       & 0.0257                     & 0.029        & 320.9317      & 9162.6501    \\
            kc1                             & 0.0053         & 0.0083       & 0.0115                     & 0.0125       & 235.6677      & 21622.0294   \\
            pc1                             & 0.0043         & 0.0056       & 0.0088                     & 0.0084       & 187.0925      & 1052.9057    \\
            satimage                        & 0.0215         & 0.0526       & 0.0815                     & 0.052        & 734.1355      & 32469.5402   \\
            spectf                          & 0.0099         & 0.0078       & 0.013                      & 0.0106       & 137.3585      & 357.6188     \\
            abalone\_17\_vs\_7\_8\_9\_10    & 0.0057         & 0.0086       & 0.0158                     & 0.0143       & 244.8049      & 2056.5642    \\
            abalone\_19\_vs\_10\_11\_12\_13 & 0.0055         & 0.0071       & 0.0118                     & 0.0109       & 170.4024      & 1448.2401    \\
            abalone\_20\_vs\_8\_9\_10       & 0.0055         & 0.0079       & 0.0147                     & 0.0119       & 183.5537      & 1594.7504    \\
            abalone\_21\_vs\_8              & 0.0042         & 0.0055       & 0.0073                     & 0.0064       & 98.9843       & 534.1806     \\
            abalone\_3\_vs\_11              & 0.0042         & 0.0046       & 0.0069                     & 0.0059       & 63.8179       & 473.0562     \\
            abalone9\_18                    & 0.0048         & 0.0061       & 0.0079                     & 0.0073       & 87.9872       & 1160.8417    \\
            car\_good                       & 0.0047         & 0.0054       & 0.0096                     & 0.0151       & 143.3904      & 6318.2594    \\
            car\_vgood                      & 0.0047         & 0.0067       & 0.0106                     & 0.0152       & 140.254       & 22054.9514   \\
            \hline 
            \bf{Average} & \bf{0.0064} &\bf{0.0102} &\bf{0.0152} &\bf{0.0136} &\bf{196.7055} &\bf{6379.0816}
        \end{longtable}
    \end{center}
    
    Table \ref{table:cWGAN_results} shows the average AUC-Scores achieved by the different methods on the first 15 datasets. Even though draGAN seems to clearly outperform the rest (especially cWGAN), we want to emphasize that using merely 15 datasets is in no way statistically significant, and the results listed here are unlikely to generalize. This is why, cWGAN has been omitted from Table \ref{table:results}. 
    
    \begin{center}
        \centering
        \tiny
        \begin{longtable}{llccccc}
            \caption{AUC-Score}
            \label{table:cWGAN_results}\\
            Dataset                                  & Vanilla           & SMOTE             & Polynom\_fit\_SMOTE        & MixUp             & draGAN (ours) & cWGAN\\
                \hline
            cm1                             & 0.729          & 0.7292       & 0.7292                     & 0.7292       & \bf{0.738}         & 0.7291       \\
            german                          & 0.5537         & 0.5537       & 0.5537                     & 0.5537       & \bf{0.661}         & 0.5537       \\
            hepatitis                       & 0.742          & 0.7428       & 0.7788                     & 0.7639       & \bf{0.783}         & 0.7454       \\
            hypothyroid                     & 0.8111         & 0.8445       & \bf{0.8483}                     & 0.826        & 0.8278        & 0.832        \\
            kc1                             & 0.7962         & 0.7962       & 0.7962                     & 0.7962       & \bf{0.7972}        & 0.7964       \\
            pc1                             & 0.657          & 0.657        & 0.657                      & 0.657        & \bf{0.7122}        & 0.5082       \\
            satimage                        & 0.7027         & 0.7057       & \bf{0.7153}                     & 0.7008       & 0.6902        & 0.7127       \\
            spectf                          & 0.8522         & 0.8501       & 0.8509                     & \bf{0.8522}       & 0.8116        & 0.8458       \\
            abalone\_17\_vs\_7\_8\_9\_10    & 0.7512         & 0.8436       & 0.8486                     & 0.7626       & \bf{0.9304}        & 0.6924       \\
            abalone\_19\_vs\_10\_11\_12\_13 & 0.454          & 0.7277       & 0.7255                     & 0.5064       & \bf{0.8063}        & 0.4596       \\
            abalone\_20\_vs\_8\_9\_10       & 0.7548         & 0.9094       & 0.9167                     & 0.7703       & \bf{0.9747}        & 0.8454       \\
            abalone\_21\_vs\_8              & 0.8957         & 0.9139       & 0.9157                     & 0.9014       & \bf{0.966}         & 0.9116       \\
            abalone\_3\_vs\_11              & \bf{1}              & \bf{1}            & \bf{1}                          & \bf{1}            & 0.9995        & \bf{1}            \\
            abalone9\_18                    & 0.731          & 0.7801       & 0.7955                     & 0.7401       & \bf{0.938}         & 0.7436       \\
            car\_good                       & \bf{0.6223}         & 0.6127       & 0.6205                     & 0.6166       & 0.5922        & 0.4932       \\
            car\_vgood                      & 0.9308         & 0.9335       & \bf{0.9343}                     & 0.9296       & 0.9252        & 0.8768  \\
            \hline 
            \bf{Average} & \bf{0.7490} &\bf{0.7875} &\bf{0.7929} &\bf{0.7566} &\bf{0.8221} &\bf{0.7341}
        \end{longtable}
    \end{center}

    \subsection{Detailed results}
    \label{appendix:detailed_results}
    
    Tables (\ref{table:results_f1_score}, \ref{table:results_g1_score}) show the performances achieved by the different benchmarking algorithms using a number of discrete measures of success. The decision boundary for each algorithm was determined using (\ref{eqn:decision_bound}).
    
    \begin{equation}
        \label{eqn:decision_bound}
        \text{Decision-Threshold} = \max_{\tiny \text{Threshold}} (\text{TPR}_{\tiny \text{Threshold}}-\text{FPR}_{\tiny \text{Threshold}})
    \end{equation}
    where TPR is (\ref{eqn:TPR}) and FPR (\ref{eqn:FPR})

    \begin{center}
    \centering
    \tiny
        \begin{longtable}{llccccc}
        \caption{F1-Score}
        \label{table:results_f1_score}\\
        Nr & Dataset                                  & Vanilla           & SMOTE             & Polynom\_fit\_SMOTE        & MixUp             & draGAN (ours) \\
            \hline
            0  & cm1                                      & \bf{0.3109}         & \bf{0.3109}       & \bf{0.3109}                     & \bf{0.3109}       & \bf{0.3064}        \\
            1  & german                                   & 0.3926         & 0.3926       & 0.3926                     & 0.3926       & \bf{0.5058}        \\
            2  & hepatitis                                & 0.461          & 0.4784       & 0.4995                     & 0.4684       & \bf{0.5168}        \\
            3  & hypothyroid                              & 0.2183         & 0.2876       & \bf{0.3086}                     & 0.2348       & 0.2434        \\
            4  & kc1                                      & \bf{0.4884}         & \bf{0.4884}       & \bf{0.4884}                     & \bf{0.4884}       & 0.4685        \\
            5  & pc1                                      & 0.254          & 0.254        & 0.254                      & 0.254        & \bf{0.2621}        \\
            6  & satimage                                 & 0.3044         & 0.3056       & \bf{0.3074}                     & 0.3071       & 0.2815        \\
            7  & spectf                                   & 0.6083         & \bf{0.6156}       & 0.6107                     & 0.6085       & 0.5601        \\
            8  & abalone\_17\_vs\_7\_8\_9\_10             & 0.1362         & 0.2496       & 0.2334                     & 0.1543       & \bf{0.3314}        \\
            9  & abalone\_19\_vs\_10\_11\_12\_13          & 0.0412         & 0.1037       & 0.124                      & 0.0492       & \bf{0.1313}        \\
            10 & abalone\_20\_vs\_8\_9\_10                & 0.0846         & 0.2288       & 0.194                      & 0.0933       & \bf{0.3356}        \\
            11 & abalone\_21\_vs\_8                       & 0.3026         & 0.3558       & 0.3558                     & 0.3431       & \bf{0.4585}        \\
            12 & abalone\_3\_vs\_11                       & \bf{0.8}            & \bf{0.8}          & \bf{0.8}                        & \bf{0.8}          & 0.7911        \\
            13 & abalone9\_18                             & 0.2685         & 0.2774       & 0.3107                     & 0.263        & \bf{0.447}         \\
            14 & car\_good                                & 0.103          & \bf{0.1141}       & 0.1086                     & 0.1079       & 0.1081        \\
            15 & car\_vgood                               & 0.3169         & 0.3368       & \bf{0.3387}                     & 0.3144       & 0.3134        \\
            16 & cleveland\_0\_vs\_4                      & 0.1694         & 0.1659       & 0.1824                     & 0.1649       & \bf{0.2147}        \\
            17 & ecoli\_0\_1\_3\_7\_vs\_2\_6              & 0.0417         & 0.0536       & 0.0536                     & 0.0369       & \bf{0.0838}        \\
            18 & ecoli\_0\_1\_4\_6\_vs\_5                 & 0.6148         & 0.6296       & 0.6348                     & \bf{0.6354}       & 0.5965        \\
            19 & ecoli\_0\_1\_4\_7\_vs\_2\_3\_5\_6        & 0.4611         & 0.4676       & 0.4583                     & \bf{0.4905}       & 0.4517        \\
            20 & ecoli\_0\_1\_4\_7\_vs\_5\_6              & \bf{0.591}          & 0.5822       & 0.5645                     & 0.5638       & 0.5673        \\
            21 & ecoli\_0\_1\_vs\_2\_3\_5                 & \bf{0.6532}         & 0.5647       & 0.6011                     & 0.5915       & 0.6152        \\
            22 & ecoli\_0\_1\_vs\_5                       & 0.6434         & 0.6512       & \bf{0.6929}                     & 0.6306       & 0.6919        \\
            23 & ecoli\_0\_2\_3\_4\_vs\_5                 & 0.6435         & 0.6343       & \bf{0.6651}                     & 0.6607       & 0.5903        \\
            24 & ecoli\_0\_2\_6\_7\_vs\_3\_5              & 0.5759         & 0.5487       & 0.5674                     & \bf{0.5776}       & 0.5015        \\
            25 & ecoli\_0\_3\_4\_6\_vs\_5                 & 0.6762         & \bf{0.6921}       & 0.6762                     & 0.673        & 0.629         \\
            26 & ecoli\_0\_3\_4\_7\_vs\_5\_6              & 0.5585         & 0.5545       & 0.5439                     & 0.5416       & \bf{0.5836}        \\
            27 & ecoli\_0\_3\_4\_vs\_5                    & 0.6268         & 0.6112       & \bf{0.6317}                     & 0.6255       & 0.6101        \\
            28 & ecoli\_0\_4\_6\_vs\_5                    & 0.6762         & 0.675        & 0.6619                     & \bf{0.6847}       & 0.6133        \\
            29 & ecoli\_0\_6\_7\_vs\_3\_5                 & 0.5779         & \bf{0.5917}       & 0.5902                     & 0.5854       & 0.553         \\
            30 & ecoli\_0\_6\_7\_vs\_5                    & \bf{0.6667}         & 0.6556       & 0.6286                     & 0.6318       & 0.5615        \\
            31 & flaref                                   & 0.2632         & 0.3118       & 0.316                      & 0.2794       & \bf{0.3198}        \\
            32 & glass\_0\_1\_4\_6\_vs\_2                 & 0.2617         & 0.2385       & \bf{0.2661}                     & 0.2518       & 0.2459        \\
            33 & glass\_0\_1\_5\_vs\_2                    & \bf{0.218}          & 0.2108       & 0.2154                     & 0.2087       & 0.1941        \\
            34 & glass\_0\_1\_6\_vs\_2                    & 0.1812         & 0.1681       & 0.1812                     & 0.1641       & \bf{0.2206}        \\
            35 & glass\_0\_1\_6\_vs\_5                    & 0.3778         & 0.3568       & 0.3778                     & 0.3763       & \bf{0.4422}        \\
            36 & glass\_0\_4\_vs\_5                       & \bf{0.4444}         & \bf{0.4444}       & \bf{0.4444}                     & \bf{0.4444}       & 0.3404        \\
            37 & glass\_0\_6\_vs\_5                       & 0.42           & 0.3844       & 0.42                       & 0.4022       & \bf{0.437}         \\
            38 & glass\_0\_1\_2\_3\_vs\_4\_5\_6           & 0.8628         & \bf{0.871}        & 0.8628                     & 0.8592       & 0.8446        \\
            39 & glass0                                   & 0.6665         & 0.6621       & 0.6665                     & 0.6641       & \bf{0.7169}        \\
            40 & glass1                                   & 0.3028         & 0.4119       & 0.3028                     & 0.39         & \bf{0.4996}        \\
            41 & glass6                                   & 0.8388         & 0.8247       & \bf{0.8388}                     & \bf{0.8388}       & 0.7578        \\
            42 & glass2                                   & 0.1899         & 0.1915       & 0.1899                     & 0.1889       & \bf{0.2363}        \\
            43 & glass4                                   & 0.4027         & 0.3971       & 0.4027                     & 0.4071       & \bf{0.4354}        \\
            44 & glass5                                   & 0.3205         & 0.3205       & 0.3205                     & 0.3229       & \bf{0.3631}        \\
            45 & kddcup\_buffer\_overflow\_vs\_back       & \bf{0.9091}         & \bf{0.9091}       & \bf{0.9091}                     & \bf{0.9091}       & \bf{0.9091}        \\
            46 & kddcup\_guess\_passwd\_vs\_satan         & \bf{0.9417}         & \bf{0.9417}       & \bf{0.9417}                     & \bf{0.9417}       & 0.9036        \\
            47 & kddcup\_land\_vs\_portsweep              & \bf{0.8421}         & \bf{0.8421}       & \bf{0.8421}                     & \bf{0.8421}       & 0.8135        \\
            48 & kddcup\_land\_vs\_satan                  & 0.8254         & 0.8421       & 0.8421                     & 0.8254       & \bf{0.8492}        \\
            49 & kr\_vs\_k\_one\_vs\_fifteen              & \bf{0.9669}         & \bf{0.9669}       & \bf{0.9669}                     & \bf{0.9669}       & \bf{0.9669}        \\
            50 & kr\_vs\_k\_three\_vs\_eleven1            & 0.9136         & 0.9504       & 0.9447                     & 0.9275       & \bf{0.9604}        \\
            51 & kr\_vs\_k\_zero\_one\_vs\_draw           & 0.7631         & 0.6925       & 0.8007                     & \bf{0.8074}       & 0.7263        \\
            52 & kr\_vs\_k\_zero\_vs\_eight               & 0.3773         & 0.3042       & 0.3053                     & \bf{0.3804}       & 0.3382        \\
            53 & kr\_vs\_k\_zero\_vs\_fifteen             & 0.8012         & \bf{0.897}        & \bf{0.897}                      & 0.8743       & 0.8691        \\
            54 & led7digit\_0\_2\_4\_5\_6\_7\_8\_9\_vs\_1 & 0.5728         & \bf{0.6948}       & 0.6131                     & 0.5674       & 0.5662        \\
            55 & lymphography\_normal\_fibrosis           & 0.0667         & \bf{0.1333}       & \bf{0.1333}                     & 0.0889       & 0.0822        \\
            56 & page\_blocks\_1\_3\_vs\_4                & 0.5245         & 0.5268       & 0.5268                     & 0.5268       & \bf{0.6998}        \\
            57 & poker\_8\_9\_vs\_51                      & \bf{0.0671}         & 0.026        & 0.0418                     & 0.0596       & 0.0497        \\
            58 & poker\_8\_9\_vs\_6                       & 0.0353         & 0.038        & 0.0305                     & 0.035        & \bf{0.0396}        \\
            59 & poker\_8\_vs\_6                          & 0.014          & 0.0159       & 0.0168                     & 0.0169       & \bf{0.0192}        \\
            60 & poker\_9\_vs\_7                          & \bf{0.0489}         & 0.0208       & 0.0174                     & 0.0444       & 0.015         \\
            61 & shuttle\_2\_vs\_5                        & 0.6984         & 0.7126       & 0.6984                     & 0.6283       & \bf{0.8097}        \\
            62 & shuttle\_6\_vs\_2\_3                     & \bf{0.6667}         & \bf{0.6667}       & \bf{0.6667}                     & \bf{0.6667}       & 0.6444        \\
            63 & shuttle\_c0\_vs\_c4                      & 0.8588         & 0.9132       & 0.9006                     & 0.8704       & \bf{0.9766}        \\
            64 & shuttle\_c2\_vs\_c4                      & 0.0444         & 0.0714       & 0.0444                     & 0.0658       & \bf{0.0714}        \\
            65 & vowel0                                   & \bf{0.8079}         & 0.7839       & 0.8002                     & 0.801        & 0.6253        \\
            66 & winequality\_red\_3\_vs\_5               & 0.1155         & \bf{0.1461}       & 0.144                      & 0.1102       & 0.0774        \\
            67 & winequality\_red\_4                      & 0.1232         & 0.1461       & \bf{0.1479}                     & 0.1221       & 0.1328        \\
            68 & winequality\_red\_8\_vs\_6               & 0.0955         & 0.1247       & \bf{0.1461}                     & 0.0977       & 0.1412        \\
            69 & winequality\_red\_8\_vs\_6\_7            & 0.0389         & 0.0505       & 0.0595                     & 0.0398       & \bf{0.1024}        \\
            70 & winequality\_white\_3\_9\_vs\_5          & 0.0753         & 0.0399       & 0.072                      & \bf{0.0949}       & 0.087         \\
            71 & winequality\_white\_3\_vs\_7             & 0.22           & 0.0759       & \bf{0.2212}                     & 0.2158       & 0.1118        \\
            72 & winequality\_white\_9\_vs\_4             & \bf{0}              & \bf{0}            & \bf{0}                          & \bf{0}            & \bf{0}             \\
            73 & yeast\_0\_2\_5\_6\_vs\_3\_7\_8\_9        & 0.5494         & 0.5648       & 0.5572                     & \bf{0.5756}       & 0.5399        \\
            74 & yeast\_0\_2\_5\_7\_9\_vs\_3\_6\_8        & 0.6839         & 0.6746       & \bf{0.6864}                     & 0.6786       & 0.6513        \\
            75 & yeast\_0\_3\_5\_9\_vs\_7\_8              & 0.3924         & 0.3623       & \bf{0.4012}                     & 0.3787       & 0.3827        \\
            76 & yeast\_0\_5\_6\_7\_9\_vs\_4              & 0.5198         & 0.514        & 0.4761                     & \bf{0.5267}       & 0.4469        \\
            77 & yeast\_1\_2\_8\_9\_vs\_7                 & 0.1382         & 0.1573       & 0.1336                     & 0.1479       & \bf{0.161}         \\
            78 & yeast\_1\_4\_5\_8\_vs\_7                 & 0.1616         & \bf{0.1782}       & 0.1613                     & 0.168        & 0.1407        \\
            79 & yeast\_1\_vs\_7                          & 0.2623         & 0.3165       & 0.3158                     & 0.274        & \bf{0.3465}        \\
            80 & yeast\_2\_vs\_4                          & 0.7065         & 0.7211       & 0.7226                     & \bf{0.7264}       & 0.5624        \\
            81 & yeast\_2\_vs\_8                          & 0.4671         & 0.4302       & \bf{0.4744}                     & 0.4692       & 0.3353        \\
            82 & yeast4                                   & 0.3358         & 0.2814       & \bf{0.3435}                     & 0.3356       & 0.2916        \\
            83 & yeast5                                   & 0.6107         & \bf{0.6145}       & 0.6028                     & 0.6089       & 0.6144        \\
            84 & yeast6                                   & \bf{0.3851}         & 0.3305       & 0.3314                     & 0.3612       & 0.3186        \\
            85 & yeast1                                   & 0.6038         & 0.6047       & \bf{0.6107}                     & 0.6088       & 0.6024        \\
            86 & yeast3                                   & 0.6531         & \bf{0.6652}       & 0.6632                     & 0.6561       & 0.6355        \\
            87 & zoo\_3                                   & \bf{0}              & \bf{0}            & \bf{0}                          & \bf{0}            & \bf{0}             \\
            88 & iris0                                    & \bf{0.9474}         & \bf{0.9474}       & \bf{0.9474}                     & \bf{0.9474}       & \bf{0.9474}        \\
            89 & new\_thyroid1                            & 0.6757         & 0.7021       & 0.7028                     & 0.6797       & \bf{0.8839}        \\
            90 & pima                                     & 0.6176         & 0.633        & 0.6473                     & 0.6223       & \bf{0.6666}        \\
            91 & segment0                                 & 0.4825         & 0.5083       & 0.5075                     & 0.4967       & \bf{0.7591}        \\
            92 & vehicle0                                 & 0.5869         & 0.5908       & 0.5907                     & 0.5897       & \bf{0.821}         \\
            93 & wisconsin                                & 0.963          & 0.9628       & 0.9611                     & \bf{0.9635}       & 0.953         \\
            \hline 
               & \bf{Average}                                      & \bf{0.4487}    & \bf{0.4549}  & \bf{0.4592}                & \bf{0.4513}   & \bf{0.4641}
        \end{longtable}
    \end{center}

    \begin{center}
    \centering
    \tiny
        \begin{longtable}{llccccc}
        \caption{G-Score}
        \label{table:results_g1_score}\\
        Nr & Dataset                                  & Vanilla           & SMOTE             & Polynom\_fit\_SMOTE        & MixUp             & draGAN (ours) \\
            \hline
        0  & cm1                                      & 0.649          & 0.649        & 0.649                      & 0.649        & \bf{0.6823}        \\
        1  & german                                   & 0.5309         & 0.5309       & 0.5309                     & 0.5309       & \bf{0.6096}        \\
        2  & hepatitis                                & 0.6586         & 0.6731       & \bf{0.6849}                     & 0.6752       & 0.6847        \\
        3  & hypothyroid                              & 0.7477         & 0.811        & \bf{0.8258}                     & 0.7703       & 0.7819        \\
        4  & kc1                                      & \bf{0.7276}         & \bf{0.7276}       & \bf{0.7276}                     & \bf{0.7276}       & 0.7255        \\
        5  & pc1                                      & 0.5741         & 0.5741       & 0.5741                     & 0.5741       & \bf{0.6799}        \\
        6  & satimage                                 & 0.6789         & 0.6796       & \bf{0.6801}                     & 0.6687       & 0.6512        \\
        7  & spectf                                   & 0.7884         & 0.7845       & 0.7825                     & \bf{0.7911}       & 0.745         \\
        8  & abalone\_17\_vs\_7\_8\_9\_10             & 0.6838         & 0.7542       & 0.7597                     & 0.6947       & \bf{0.8562}        \\
        9  & abalone\_19\_vs\_10\_11\_12\_13          & 0.4094         & 0.6466       & 0.6119                     & 0.4439       & \bf{0.7265}        \\
        10 & abalone\_20\_vs\_8\_9\_10                & 0.6349         & 0.7873       & 0.7958                     & 0.643        & \bf{0.8612}        \\
        11 & abalone\_21\_vs\_8                       & 0.7064         & 0.7207       & 0.7208                     & 0.7127       & \bf{0.7556}        \\
        12 & abalone\_3\_vs\_11                       & \bf{0.8165}         & \bf{0.8165}       & \bf{0.8165}                     & \bf{0.8165}       & 0.8162        \\
        13 & abalone9\_18                             & 0.6734         & 0.7032       & 0.7145                     & 0.6606       & \bf{0.8478}        \\
        14 & car\_good                                & 0.5454         & \bf{0.6064}       & 0.579                      & 0.5754       & 0.5883        \\
        15 & car\_vgood                               & 0.8758         & \bf{0.8829}       & 0.8827                     & 0.8742       & 0.8687        \\
        16 & cleveland\_0\_vs\_4                      & 0.4359         & 0.4181       & 0.4429                     & 0.4529       & \bf{0.4663}        \\
        17 & ecoli\_0\_1\_3\_7\_vs\_2\_6              & 0.2361         & 0.2498       & 0.2498                     & 0.2215       & \bf{0.2601}        \\
        18 & ecoli\_0\_1\_4\_6\_vs\_5                 & \bf{0.8008}         & 0.7905       & 0.7867                     & 0.7989       & 0.7722        \\
        19 & ecoli\_0\_1\_4\_7\_vs\_2\_3\_5\_6        & 0.741          & \bf{0.7499}       & 0.7208                     & 0.747        & 0.7375        \\
        20 & ecoli\_0\_1\_4\_7\_vs\_5\_6              & 0.7646         & 0.7649       & 0.7598                     & \bf{0.7693}       & 0.7612        \\
        21 & ecoli\_0\_1\_vs\_2\_3\_5                 & \bf{0.8048}         & 0.7754       & 0.7866                     & 0.7882       & 0.7678        \\
        22 & ecoli\_0\_1\_vs\_5                       & \bf{0.8266}         & 0.8154       & 0.8091                     & 0.8252       & 0.7886        \\
        23 & ecoli\_0\_2\_3\_4\_vs\_5                 & \bf{0.7959}         & 0.7771       & 0.784                      & 0.7922       & 0.7491        \\
        24 & ecoli\_0\_2\_6\_7\_vs\_3\_5              & 0.7412         & 0.7219       & 0.7356                     & \bf{0.7414}       & 0.6922        \\
        25 & ecoli\_0\_3\_4\_6\_vs\_5                 & 0.7865         & \bf{0.7875}       & 0.7865                     & 0.7858       & 0.7715        \\
        26 & ecoli\_0\_3\_4\_7\_vs\_5\_6              & 0.7692         & \bf{0.7694}       & 0.7672                     & 0.7623       & 0.7635        \\
        27 & ecoli\_0\_3\_4\_vs\_5                    & \bf{0.7878}         & 0.7686       & 0.7676                     & 0.7868       & 0.728         \\
        28 & ecoli\_0\_4\_6\_vs\_5                    & 0.786          & 0.782        & 0.7807                     & \bf{0.7865}       & 0.7666        \\
        29 & ecoli\_0\_6\_7\_vs\_3\_5                 & 0.7553         & 0.7422       & 0.7513                     & \bf{0.757}        & 0.7243        \\
        30 & ecoli\_0\_6\_7\_vs\_5                    & \bf{0.8105}         & 0.8096       & 0.8038                     & 0.8032       & 0.7553        \\
        31 & flaref                                   & 0.8076         & \bf{0.8302}       & 0.8295                     & 0.8203       & 0.8133        \\
        32 & glass\_0\_1\_4\_6\_vs\_2                 & 0.5591         & 0.5352       & \bf{0.5608}                     & \bf{0.5608}       & 0.5438        \\
        33 & glass\_0\_1\_5\_vs\_2                    & \bf{0.5752}         & 0.5599       & 0.5714                     & 0.5559       & 0.5139        \\
        34 & glass\_0\_1\_6\_vs\_2                    & 0.4916         & 0.4678       & 0.4916                     & 0.4655       & \bf{0.5713}        \\
        35 & glass\_0\_1\_6\_vs\_5                    & 0.5468         & 0.5403       & 0.5468                     & 0.5461       & \bf{0.5589}        \\
        36 & glass\_0\_4\_vs\_5                       & \bf{0.5361}         & \bf{0.5361}       & \bf{0.5361}                     & \bf{0.5361}       & 0.4507        \\
        37 & glass\_0\_6\_vs\_5                       & 0.4701         & 0.4372       & 0.4701                     & 0.4537       & \bf{0.5465}        \\
        38 & glass\_0\_1\_2\_3\_vs\_4\_5\_6           & 0.9054         & 0.9043       & \bf{0.9054}                     & 0.9032       & 0.9047        \\
        39 & glass0                                   & 0.693          & 0.7117       & 0.693                      & 0.6976       & \bf{0.7893}        \\
        40 & glass1                                   & 0.3529         & 0.4859       & 0.3529                     & 0.4319       & \bf{0.5199}        \\
        41 & glass6                                   & 0.8638         & \bf{0.8651}       & 0.8638                     & 0.8638       & 0.861         \\
        42 & glass2                                   & 0.5716         & 0.5532       & 0.5716                     & 0.5639       & \bf{0.6128}        \\
        43 & glass4                                   & 0.6496         & 0.6501       & 0.6496                     & 0.6501       & \bf{0.6677}        \\
        44 & glass5                                   & 0.5462         & 0.5462       & 0.5462                     & \bf{0.5468}       & 0.5419        \\
        45 & kddcup\_buffer\_overflow\_vs\_back       & \bf{0.9129}         & \bf{0.9129}       & \bf{0.9129}                     & \bf{0.9129}       & \bf{0.9129}        \\
        46 & kddcup\_guess\_passwd\_vs\_satan         & \bf{0.9513}         & \bf{0.9513}       & \bf{0.9513}                     & \bf{0.9513}       & 0.9493        \\
        47 & kddcup\_land\_vs\_portsweep              & \bf{0.8713}         & \bf{0.8713}       & \bf{0.8713}                     & \bf{0.8713}       & 0.8686        \\
        48 & kddcup\_land\_vs\_satan                  & 0.8712         & 0.8714       & 0.8714                     & 0.8712       & \bf{0.8715}        \\
        49 & kr\_vs\_k\_one\_vs\_fifteen              & \bf{0.9674}         & \bf{0.9674}       & \bf{0.9674}                     & \bf{0.9674}       & \bf{0.9674}        \\
        50 & kr\_vs\_k\_three\_vs\_eleven1            & 0.9669         & 0.9681       & 0.9679                     & 0.9674       & \bf{0.9684}        \\
        51 & kr\_vs\_k\_zero\_one\_vs\_draw           & 0.9487         & 0.9482       & 0.9524                     & \bf{0.9528}       & 0.9517        \\
        52 & kr\_vs\_k\_zero\_vs\_eight               & 0.82           & \bf{0.8456}       & 0.8431                     & 0.8251       & 0.8414        \\
        53 & kr\_vs\_k\_zero\_vs\_fifteen             & 0.9004         & \bf{0.9018}       & \bf{0.9018}                     & 0.9015       & 0.9013        \\
        54 & led7digit\_0\_2\_4\_5\_6\_7\_8\_9\_vs\_1 & 0.7302         & \bf{0.8183}       & 0.721                      & 0.7343       & 0.7184        \\
        55 & lymphography\_normal\_fibrosis           & 0.1336         & \bf{0.1414}       & \bf{0.1414}                     & 0.1371       & 0.1363        \\
        56 & page\_blocks\_1\_3\_vs\_4                & 0.8374         & 0.8385       & 0.8385                     & 0.8385       & \bf{0.8627}        \\
        57 & poker\_8\_9\_vs\_51                      & 0.4602         & 0.3696       & 0.3968                     & \bf{0.4605}       & 0.4188        \\
        58 & poker\_8\_9\_vs\_6                       & 0.3894         & \bf{0.4065}       & 0.3121                     & 0.3671       & 0.3249        \\
        59 & poker\_8\_vs\_6                          & 0.2266         & 0.2201       & 0.1994                     & 0.2821       & \bf{0.3313}        \\
        60 & poker\_9\_vs\_7                          & \bf{0.3012}         & 0.1142       & 0.1072                     & 0.2751       & 0.0878        \\
        61 & shuttle\_2\_vs\_5                        & 0.9426         & \bf{0.9429}       & 0.9426                     & 0.9375       & 0.9414        \\
        62 & shuttle\_6\_vs\_2\_3                     & \bf{0.7071}         & \bf{0.7071}       & \bf{0.7071}                     & \bf{0.7071}       & 0.7055        \\
        63 & shuttle\_c0\_vs\_c4                      & 0.9654         & 0.9705       & 0.9694                     & 0.9664       & \bf{0.9793}        \\
        64 & shuttle\_c2\_vs\_c4                      & 0.1225         & \bf{0.1265}       & 0.1225                     & 0.1204       & 0.1251        \\
        65 & vowel0                                   & \bf{0.9161}         & 0.9128       & 0.915                      & 0.9151       & 0.8893        \\
        66 & winequality\_red\_3\_vs\_5               & \bf{0.5795}         & 0.5773       & 0.5762                     & 0.4836       & 0.5329        \\
        67 & winequality\_red\_4                      & 0.5654         & 0.5584       & 0.5595                     & 0.5645       & \bf{0.6047}        \\
        68 & winequality\_red\_8\_vs\_6               & 0.5785         & 0.5893       & 0.5919                     & 0.5783       & \bf{0.6823}        \\
        69 & winequality\_red\_8\_vs\_6\_7            & 0.3558         & 0.3465       & 0.3509                     & 0.3572       & \bf{0.6229}        \\
        70 & winequality\_white\_3\_9\_vs\_5          & 0.1772         & 0.2109       & 0.1076                     & 0.2187       & \bf{0.5852}        \\
        71 & winequality\_white\_3\_vs\_7             & \bf{0.4691}         & 0.4363       & 0.4689                     & 0.4682       & 0.3758        \\
        72 & winequality\_white\_9\_vs\_4             & \bf{0}              & \bf{0}            & \bf{0}                          & \bf{0}            & \bf{0}             \\
        73 & yeast\_0\_2\_5\_6\_vs\_3\_7\_8\_9        & \bf{0.7825}         & 0.7607       & 0.7812                     & 0.7816       & 0.7462        \\
        74 & yeast\_0\_2\_5\_7\_9\_vs\_3\_6\_8        & \bf{0.8358}         & 0.8262       & 0.8346                     & 0.8296       & 0.8053        \\
        75 & yeast\_0\_3\_5\_9\_vs\_7\_8              & 0.679          & 0.677        & \bf{0.6865}                     & 0.6808       & 0.678         \\
        76 & yeast\_0\_5\_6\_7\_9\_vs\_4              & 0.778          & \bf{0.7857}       & 0.7789                     & 0.7723       & 0.7609        \\
        77 & yeast\_1\_2\_8\_9\_vs\_7                 & 0.627          & 0.6626       & 0.6569                     & 0.6285       & \bf{0.6695}        \\
        78 & yeast\_1\_4\_5\_8\_vs\_7                 & 0.6            & 0.5979       & 0.5917                     & \bf{0.6082}       & 0.5714        \\
        79 & yeast\_1\_vs\_7                          & 0.6633         & 0.7246       & 0.7119                     & 0.6947       & \bf{0.7263}        \\
        80 & yeast\_2\_vs\_4                          & 0.8516         & \bf{0.862}        & 0.8599                     & 0.8516       & 0.8286        \\
        81 & yeast\_2\_vs\_8                          & 0.6396         & 0.6456       & 0.6404                     & \bf{0.6481}       & 0.6474        \\
        82 & yeast4                                   & 0.8139         & 0.8087       & \bf{0.816}                      & 0.8136       & 0.8132        \\
        83 & yeast5                                   & 0.9251         & \bf{0.9256}       & 0.9252                     & 0.9251       & 0.9246        \\
        84 & yeast6                                   & 0.8281         & 0.8305       & \bf{0.834}                      & 0.8333       & 0.8287        \\
        85 & yeast1                                   & 0.7127         & 0.7144       & 0.7152                     & \bf{0.7175}       & 0.7138        \\
        86 & yeast3                                   & 0.9058         & \bf{0.9106}       & 0.9086                     & 0.9063       & 0.8733        \\
        87 & zoo\_3                                   & \bf{0}              & \bf{0}            & \bf{0}                          & \bf{0}            & \bf{0}             \\
        88 & iris0                                    & 0.9487         & 0.9487       & 0.9487                     & 0.9487       & \bf{0.9487}        \\
        89 & new\_thyroid1                            & 0.7997         & 0.8198       & 0.8209                     & 0.8104       & \bf{0.9161}        \\
        90 & pima                                     & 0.6916         & 0.7066       & 0.7194                     & 0.6931       & \bf{0.7318}        \\
        91 & segment0                                 & 0.7878         & 0.8106       & 0.8052                     & 0.7949       & \bf{0.9208}        \\
        92 & vehicle0                                 & 0.7537         & 0.7587       & 0.7579                     & 0.7576       & \bf{0.9108}        \\
        93 & wisconsin                                & \bf{0.9739}         & 0.9733       & 0.9728                     & 0.9736       & 0.9669        \\
        \hline 
         & \bf{Average} & \bf{0.6742} &\bf{0.6805} &\bf{0.6776} &\bf{0.6758} &\bf{0.696} 
        \end{longtable}
    \end{center}
    
\end{document}